\documentclass[10pt,journal]{IEEEtran}

\usepackage{amsmath,amsfonts}
\usepackage{algorithm2e}
\usepackage{array}
\usepackage{caption}
\usepackage{subcaption}
\usepackage{cite}
\usepackage{textcomp}
\usepackage{stfloats}
\usepackage{url}
\usepackage{hyperref}
\usepackage{verbatim}
\usepackage{graphicx}
\usepackage{multirow}
\hypersetup{urlcolor=cyan}

\graphicspath{{fig/}}
\DeclareGraphicsExtensions{.png,.jpg,.jpeg}

\title{FM-Fusion: Instance-aware Semantic Mapping \\Boosted by Vision-Language Foundation Models}

\author{Chuhao Liu$^{1}$, Ke Wang${^{2,\ast}}$, Jieqi Shi$^{1}$, Zhijian Qiao$^{1}$ and Shaojie Shen$^{1}$ 
\thanks{Manuscript received: October 24, 2023; Accepted: January, 1, 2024.} 
\thanks{This paper was recommended for publication by Editor Markus Vincze upon evaluation of the Associate Editor and Reviewers' comments.} 
\thanks{$^{1}$Authors are with the Department of Electronic and Computer Engineering, the Hong Kong University of Science and Technology, Hong Kong, China. {\tt \footnotesize \{cliuci,jshias,zqiaoac\}@connect.ust.hk, eeshaojie@ust.hk}}
\thanks{$^{2}$Author is with the Department of Information Engineering, Chang'an University, China. {\tt \footnotesize kwangdd@chd.edu.cn}}
\thanks{$^\ast$ Corresponding author.}
}

\IEEEoverridecommandlockouts                 

\begin{document}
\maketitle

\begin{abstract}
Semantic mapping based on the supervised object detectors is sensitive to image distribution. In real-world environments, the object detection and segmentation performance can lead to a major drop,  preventing the use of semantic mapping in a wider domain. On the other hand, the development of vision-language foundation models demonstrates a strong zero-shot transferability across data distribution. It provides an opportunity to construct generalizable instance-aware semantic maps.
Hence, this work explores how to boost instance-aware semantic mapping from object detection generated from foundation models. We propose a probabilistic label fusion method to predict close-set semantic classes from open-set label measurements. An instance refinement module merges the over-segmented instances caused by inconsistent segmentation. We integrate all the modules into a unified semantic mapping system. Reading a sequence of RGB-D input, our work incrementally reconstructs an instance-aware semantic map. We evaluate the zero-shot performance of our method in ScanNet and SceneNN datasets. Our method achieves 40.3 mean average precision (mAP) on the ScanNet semantic instance segmentation task. It outperforms the traditional semantic mapping method significantly.
Code is available at \url{https://github.com/HKUST-Aerial-Robotics/FM-Fusion}.

\end{abstract}

\section{Introduction}
{I}{stance-aware} semantic mapping in indoor environments is a key module for an autonomous system to achieve a higher level of intelligence. Based on the semantic map, a mobile robot can detect loop more robust\cite{Lin2021topobj} and efficiently\cite{hughes2022hydra}. The current methods rely on supervised object detectors like Mask R-CNN \cite{he2017mask} to detect semantic instances and fuse them into an instance-level semantic map. However, the supervised object detectors are trained in specific data distribution and lack generalization ability. In deploying them in other real-world scenarios without fine-tune the networks, their performance is seriously degenerated. As a result, the reconstructed semantic map is also of poor quality in the target environment.

On the other hand, foundation models have been developing rapidly in vision-language modality \cite{kirillov2023sam} \cite{radford2021clip}. Multiple foundation models are combined to detect and segment objects. 
GroundingDINO\cite{liu2023grounding}, the latest State-of-the-Arts (SOTA) open-set object detection network, reads a text prompt and performs vision-language modal fusion. It detects objects with bounding boxes and open-set labels. The open-set labels are open vocabulary semantic classes. GroundingDINO has achieved 52.5 mAP on the zero-shot COCO object detection benchmark. It is higher than most of the supervised object detectors.
Moreover, the image tagging model recognizes anything (RAM) \cite{zhang2023ram} predicts semantic tags from an image. The tags can be encoded as a text prompt and sent to GroundingDINO. Vision foundation model segment anything (SAM)\cite{kirillov2023sam} generates precise zero-shot image segmentation results from geometric prompts, including a bounding box prompt. SAM can generate high-quality masks for detection results from GroundingDINO.

\begin{figure}[t]
	\centering	\includegraphics[width=\columnwidth]{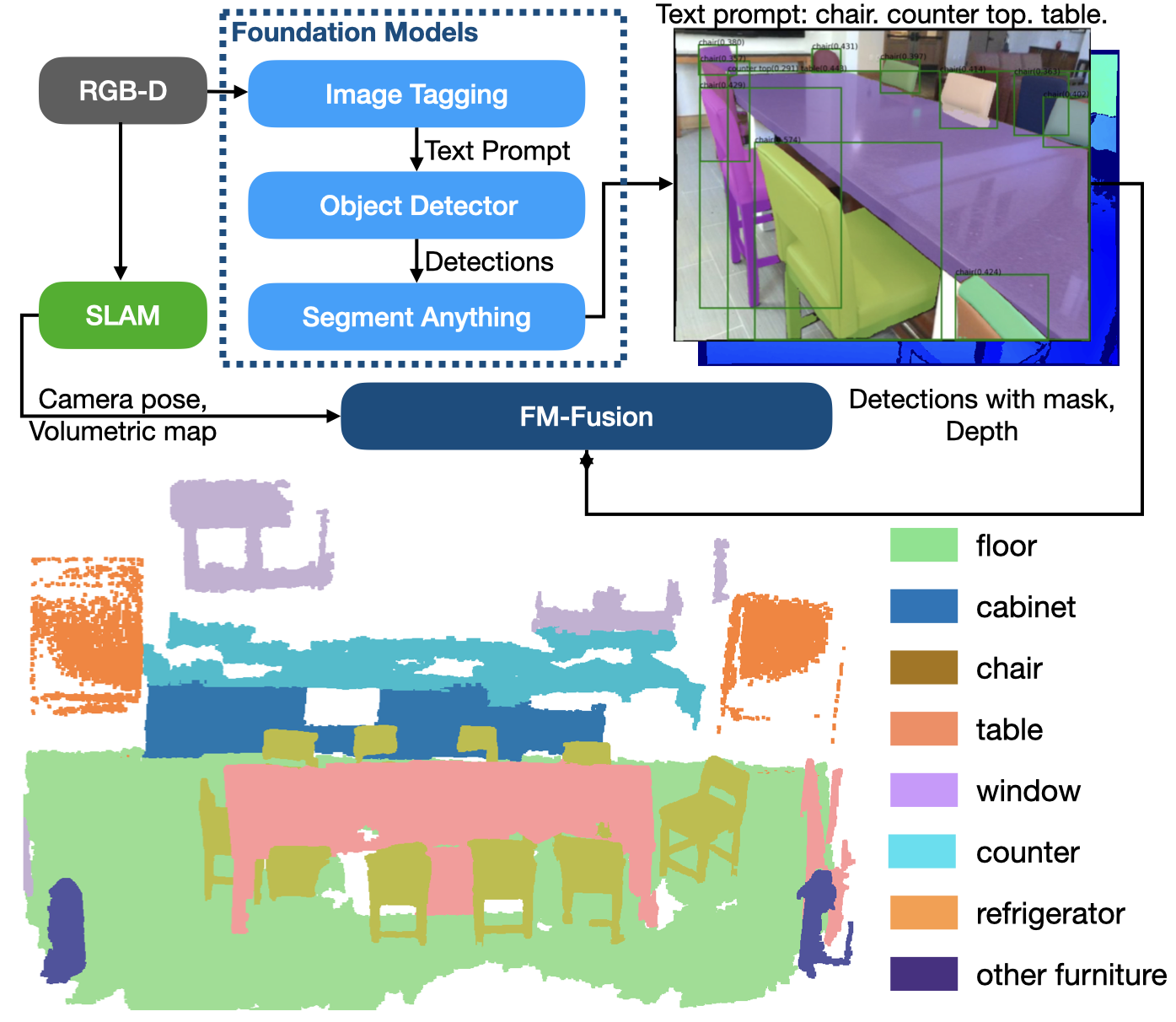}
	\caption{Our system reads a sequence of RGB-D frames. The vision-language foundation models detect objects in open-set labels and high-quality masks. The SLAM modules generate a camera pose and a global volumetric map. Our method incrementally fuses the object detections from foundation models into an instance-aware semantic map. A reconstructed semantic map from ScanNet \textit{scene0011\_01} is shown. 
}
	\label{fig-system}
	\vspace{-1.0cm}
\end{figure}

RAM, GroundingDINO, and SAM can be combined to detect objects in open-set labels and high-quality masks. All of these foundation models are trained using large-scale data and demonstrate strong zero-shot generalization ability in various image distributions. They provide a new approach for the autonomous system to reconstruct a generalizable instance-aware semantic map. This paper explores how to fuse object detection from foundation models into an instance-aware semantic map. 

To fuse object detection from foundation models, two challenges should be addressed. Firstly, the foundation models generate open-set tags or labels. However, the semantic mapping task requires each constructed instance to be classified in close-set semantic classes. A label fusion method is required to predict an instance's semantic class from a sequence of observed open-set labels. 
Secondly, SAM is operating on a single image. In dense indoor environments, SAM frequently generates inconsistent instance masks at changed viewpoints. It results in over-segmented and noisy instance volumes. Refining instance volumes integrated from inconsistent instance segmentation results is the challenge.
However, these challenges have not been considered in traditional semantic mapping works. If foundation models are directly used in a traditional semantic mapping system, they reconstruct semantic instances in a less satisfied quality.

To address such challenges, we propose a probabilistic label fusion method following the Bayes filter algorithm. Meanwhile, we refine the instance volume via merging over-segmentation and fuse instance volume with the global volumetric map. The label fusion and instance refinement modules are incrementally run in our system. 
As shown in Figure \ref{fig-system}, reading a sequence of RGB-D frames, FM-Fusion fuses the detections from foundation models and runs simultaneously with a traditional SLAM system.
Our main contributions are:
\begin{itemize}
	\item An approach to fuse the object detections from vision-language foundation models into an instance-aware semantic map. The foundation models are used without fine-tune.
	\item A probabilistic label fusion method that predicts close-set semantic classes from open-set label measurements. 
	\item Instances are refined to address inconsistent masks at changed viewpoints.
	\item The method is zero-shot evaluated in ScanNet\cite{dai2017scannet}. It outperforms the traditional semantic mapping method significantly. We further evaluate it in SceneNN \cite{hua2016scenenn} to demonstrate its robustness in other image distributions. 
\end{itemize}

\section{Related Works}
\subsection{Vision-Language Foundation Models}
The image tagging foundation model RAM \cite{zhang2023ram}, recognizes the semantic categories in the image and generates related tags. The open-set object detector, such as GLIP \cite{li2022glip} and GroundingDINO \cite{liu2023grounding}, reads a text prompt to detect the objects. The text prompt can be a sentence or a series of semantic labels. It extracts the regional image embeddings and matches the image embedding to the phrase of the text prompt through a grounding scheme. The network is trained using contrastive learning to align the image embeddings and text embeddings. The detection results contain a bounding box and a set of open-set label measurements. SAM\cite{kirillov2023sam} can precisely segment any object with a geometric prompt. It is trained with 11M images and evaluated in zero-shot benchmarks. SAM demonstrates strong generalization ability across data distribution without fine-tune. 
The combined foundation models read an image and detect objects with open-set labels and masks. We denote them as RAM-Grounded-SAM.
\footnote{https://github.com/IDEA-Research/Grounded-Segment-Anything}.

The foundation models have been applied in a series of downstream tasks without fine-tuning. Without semantic prediction, SAM3D \cite{zhang2023sam3d} projects the image-wise segmentation from SAM to a 3D point cloud map. It further merges the segments with geometric segments generated from graph-based segmentation \cite{felzenszwalb2004graphseg}. SAM is also combined with a neural radiance field to generate a novel view of objects \cite{shen2023anything3d}. On the other hand, combining the SAM or other foundation models with semantic mapping is still an open area.

\subsection{Semantic Mapping}

SemanticFusion \cite{mccormac2017semanticfusion} is a pioneer work in semantic mapping. It trains a lightweight CNN-based semantic segmentation network \cite{noh2015learning} on the NYUv2 dataset. SemanticFusion incrementally fuses the semantic labels, ignoring the instance-level information, into each surfel of the global volumetric map. In Bayesian fusing the label measurement, the semantic probability is directly provided by the object detector.
Relying on a pre-trained Mask R-CNN on the COCO dataset, Kimera \cite{Kimera2020Rosinol} uses similar methods to fuse semantic labels into a voxel map. It clusters the nearby voxels with identical semantic labels into instances. Kimera further constructs a scene graph, which is a hierarchical map representation. Based on Kimera, Hydra \cite{hughes2022hydra} utilizes the scene graph to detect loops more efficiently.

On the other hand, Fusion++ \cite{fusion++2018McCormac} directly detects semantic instances on images and fuses them into instance-wise volumetric maps. It further demonstrates that semantic landmarks can be used in loop detection. Later methods use similar methods to construct semantic instance maps but utilize the semantic landmarks in novel methods to detect loops \cite{Lin2021topobj}\cite{yu2022semanticloop}. 

Rather than a pure dense map such as a surfel map or voxel map, Voxblox++ \cite{voxblox++2019Grinvald} first generates geometric segments on each depth frame\cite{furrer2018incremental}. If the object detection masks the complete region of an instance, it can merge those broken segments generated from geometric segmentation. Then, the merged segments with their labels are fused into a global segment map through a data association strategy.

The main limitation of the current semantic mapping methods is the lack of ability to generalize. The supervised object detection networks are trained with limited source data. Considering the majority of target SLAM scenarios do not provide annotated semantic data, object detection can not be fine-tuned on the target distribution. To avoid the issue of generalization, Kimera has to experiment in a synthetic dataset\cite{Kimera2020Rosinol}, including some experiments that rely on ground-truth segmentation. Lin etc. \cite{Lin2021topobj} sets up an environment with sparsely distributed objects to reconstruct a semantic map. 
Voxblox++ evaluates a few of the 9 semantic classes in 10 scans.
Although they propose novel semantic SLAM methods, the semantic mapping module prevents their methods from being used in other real-world scenes.

To enhance robustness in the distribution shift, our method fuses the object detections from foundation models to reconstruct the instance-aware semantic map. We evaluate its zero-shot performance on the ScanNet semantic instance segmentation benchmark. It involves 20 classes in the NYUv2 label-set and evaluates their average precision(AP) in 3D space. We also show the qualitative results in several SceneNN scans, which have been used by the previous semantic mapping works.

\section{Fuse Multi-frame Detections}
 \begin{figure}[h]
 	\centering
	\vspace{-0.3cm}
 	\includegraphics[width=\columnwidth]{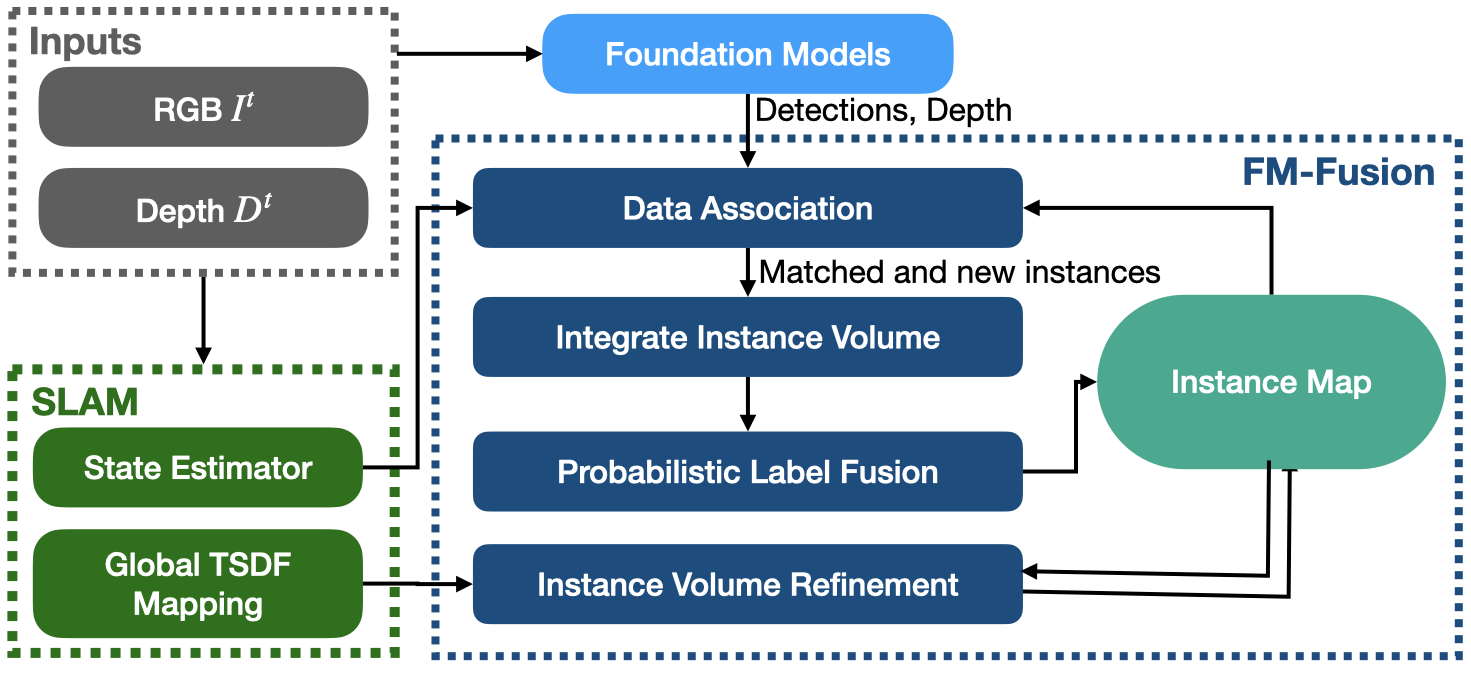}
 	\caption{System overview of FM-Fusion}
 	\label{fig-pipelines}
	\vspace{-0.6cm}
 	
 \end{figure}
 
 \subsection{Overview}
As shown in Figure \ref{fig-pipelines}, FM-Fusion reads an RGB-D sequence and reconstructs a semantic instance map. Each semantic instance is represented as $\mathbf{f}=\{L_s, \mathbf{v}\}$, where $L_s$ is its predicted semantic class and $\mathbf{v}$ is its voxel grid map. $L_s$ is predicted as a label $c_n$ over the NYUv2 label-set $\mathcal{L}_c$.
At each RGB-D frame $\{I^t, D^t\}$ at frame index $t$, RAM generates a set of possible object tags. The valid tags are encoded into the text prompt $q^t$. GroundingDINO generates object detections with each of the detection $z^t_k=\{y_i,s_i,q^t\}_i$, where $y_i$ is the predicted open-set label, $s_i$ is the corresponding similarity score and $q^t$ is the frame-wise text prompt. For each $z^t_k$, SAM generates an object mask $m_k$. 

\subsection{Prepare the object detector}\label{sec-preparedetector}
We first construct open-set labels of our interests $\mathcal{L}_o$. RAM generates various tags. Many of them are not correlated with the pre-defined labels $\mathcal{L}_c$. The labels of interest can be selected by sampling a histogram of measured labels for each semantic class in $\mathbf{L}_c$. In the ScanNet experiment, we select $38$ open-set labels to construct $\mathcal{L}_o$. 
Only the tags belonging to $\mathcal{L}_o$ are encoded into the $q^t$ and sent to GroundingDINO. GroundingDINO matches each detected object with the tags in the text prompt. The tags in $q^t$ and label measurements $\{y_i\}_i$ in each $z^t_k$ are all from the label-set $\mathcal{L}_o$.

In a single image frame, RAM can miss some objects in its generated tags due to occlusion. The missing tags further cause GroundingDINO to detect objects incorrectly. It is a natural limitation of running foundation models on a single image. To address it, we encode the detected labels in adjacent frames into the text prompt. The augmented text prompt $q^t=\bar{q^t} \cup {U}^t$, where $\bar{q^t}$ is the valid tags from RAM and $U^t$ is a set of measured labels in previous adjacent frames.
All the tags in $\bar{q^t}$ and labels in $U^t$ belong to the $\mathcal{L}_o$. The text prompt augmentation can reduce the missing tags generated from a single image. More complete tags improve the detection performance of GroundingDINO.

\subsection{Data association and integration}
In our system, each instance maintains an individual voxel grid map $\mathbf{v}$, similar to Fusion++ \cite{fusion++2018McCormac}.
Meanwhile, the SLAM module integrates a global TSDF map \cite{voxblox2017Oleynikova} separately. The advantage of separating semantic mapping and global volumetric mapping is that false or missed object detection can not affect the global volumetric map. So, in each RGB-D frame, all the observed sub-volumes are integrated into the global TSDF map despite the detection variances.

In each detection frame, data association is conducted between detection results and volumes of the existing instances. Specifically, the observed instance voxels are first queried. They can be searched by projecting the depth image into the voxel grid map of all the instances. If an instance is observed, its voxels are projected to the current RGB frame. 
For a detection $z^t_k$ and a projected instance $\mathbf{f}_j$, their intersection over union (IoU) can be calculated $\Omega(z^t_k,\mathbf{f_j})=\frac{m_k \cap r_j}{m_k \cup r_j}$, where $m_k$ is a detection mask and $r_j$ is the projected mask of an existed instance. If $\Omega(z^t_k,\mathbf{f}_j)$ is larger than a threshold, the detection $k$ is associated with instance $j$.

After data association, we integrate the voxel grid map of matched instances accordingly. Those unmatched detections initiate new instances. 
An instance voxel grid map $\mathbf{v}$ is integrated using the traditional voxel map fusion method \cite{voxblox2017Oleynikova}. Specifically, we raycast the masked depth of a detected object and update all of its observed voxels.

\subsection{Probabilistic label fusion}\label{sec:fusion}
\begin{figure}[ht]
	\centering
	\includegraphics[width=0.9\columnwidth]{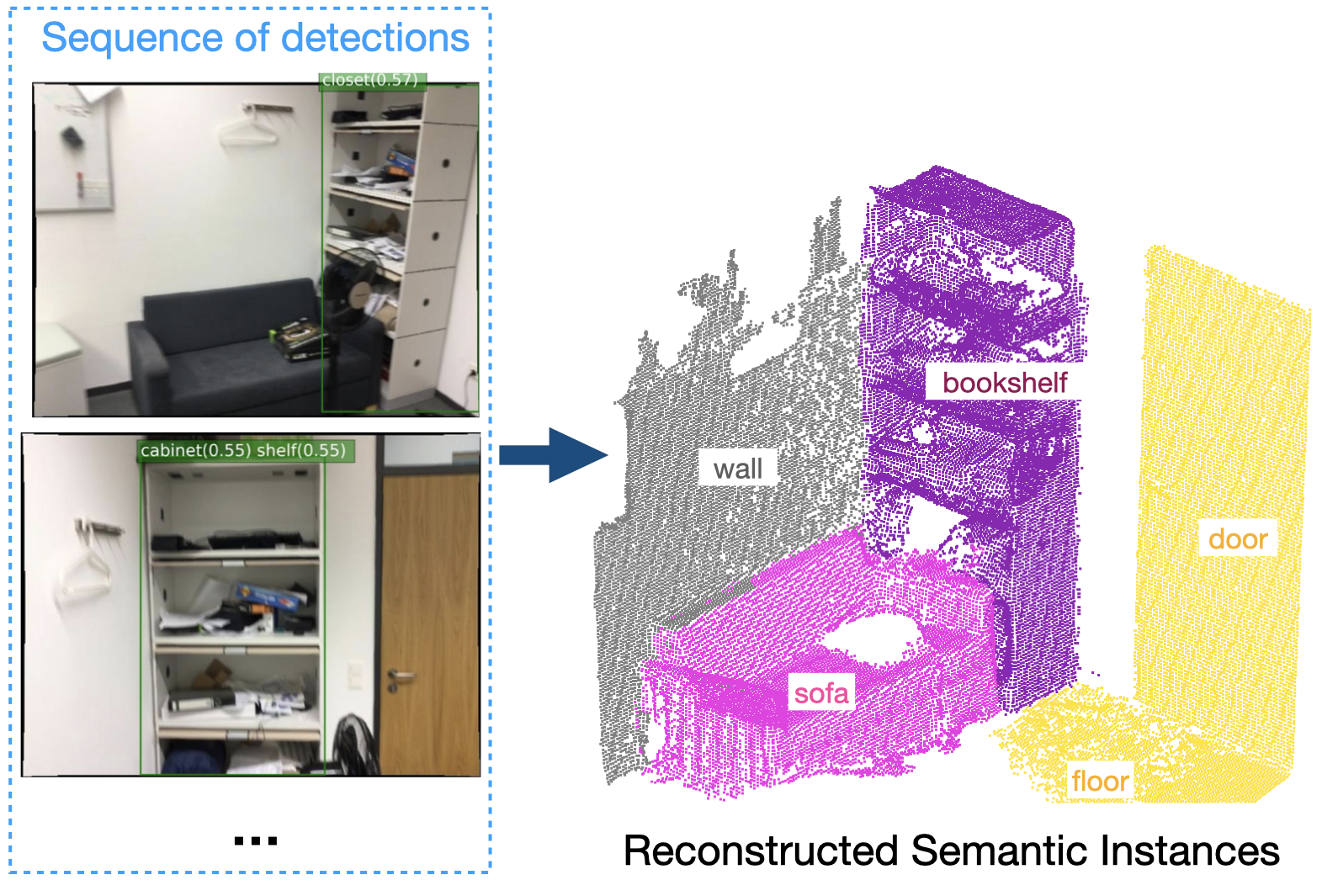}
	\vspace{-0.2cm}
	\caption{GroundingDINO detects a \emph{bookshelf} and generates multiple open-set label measurements across frames. Our label fusion module predicts its semantic class in NYUv2 label-set $\mathcal{L}_c$ from label measurements in $\mathcal{L}_o$.}
	\label{fig-framelabels}
	\vspace{-0.3cm}
\end{figure}

As shown in Figure \ref{fig-framelabels}, an object is observed by GroundingDINO across frames. Each generated detection result $z^t_k=\{y_i,s_i,q^t\}_i$ contains multiple label measurements $y_i$, the corresponding similarity score $s_i$ and a text prompt $q^t$, where $y_i=o_m, o_m \in \mathcal{L}_o$. Based on the associated detections, we predict a probability distribution $p(L_s^t=c_n)$, where $c_n \in \mathcal{L}_c$ and $t$ is the index of the image frame.

We follow the Bayes filter algorithm \cite{thrun2002probabilistic} to fuse open-set label measurements and propagate them along the image sequence.
The input to the Bayesian label fusion is detection result $z^t_k$, semantic probability distribution at the last frame $p(L_s^{t-1})$, and a uniform control input $u^t$. And it predicts the latest semantic probability distribution $p(L_s^t)$. 

\RestyleAlgo{ruled} 
\SetKwComment{Comment}{/* }{ */}
\begin{algorithm}[h]
	\caption{Bayes Filter for Label Fusion}\label{alg-bayes}
	\KwIn{$p(L_s^{t-1})$ , $z_k^t=\{y_i,s_i,q^t\}_{i\in[0:J)}$, $u^t=1$
	}
	\KwOut{$p(L_s^t)$}

	\For{$c_n \in \mathcal{L}_c$}{
	Prediction:
	\begin{equation}
	\bar{p}(L_s^t=c_n)=p(L_s^t=c_n|L_s^{t-1}=c_n,u^t)p(L_s^{t-1}=c_n)
	\end{equation}
	\begin{equation}
	\bar{p}(L_s^t=c_n)=p(L_s^{t-1}=c_n)
	\end{equation}

	Measurement Update:
	\begin{equation}\label{eq-update}
		p(L_s^t=c_n) = \eta p(z^t_k|L_s^t=c_n)\bar{p}(L_s^t=c_n)
	\end{equation}	
	\vspace{-0.8cm}
	
	\begin{equation}
		\begin{split}
			p(L_s^t&=c_n) \\
			&=\eta \Pi_{i=0}^{J-1}p(y_i,s_i,q^t|L^t_s=c_n)\bar{p}(L^t_s=c_n)
		\end{split}
	\end{equation}
	\vspace{-0.8cm}

	\begin{equation}\label{eq-scorelike}
		\begin{split}
		p(y_i&,s_i,q^t|L^t_s=c_n)\\
		&=p(s_i|y_i,q^t,L_s^t=c_n)p(y_i,q^t|L_s^t=c_n)
		\end{split}
	\end{equation}
	\vspace{-0.8cm}

	\begin{equation}\label{eq-detprompt}
			p(y_i,q^t|L_s^t=c_n)= p(y_i=o_m,\exists o_m \in q^t|L^t_s=c_n)
	\end{equation}				
		
	}
\end{algorithm}

The key part in our Bayesian label fusion module is the likelihood function $p(y_i,s_i,q^t|L_s^t=c_n)$, as shown in equation (\ref{eq-scorelike}). The score likelihood $p(s_i|y_i,q^t,L^t_s=c_n)$ is given by GroundingDINO, while label likelihood $p(y_i,q^t|L^t_s=c_n)$ should be statistic summarized. Since GroundingDINO can only detect a label $y_i$ if it is given in the text prompt, the label likelihood can be transmitted as equation (\ref{eq-detprompt}). $\exists o_m \in q^t$ denotes the detected label $o_m$ exists in the text prompt $q^t$.

Here, we further expand the label likelihood in equation (\ref{eq-detprompt}) into two conditional probabilities, 

\begin{equation} \label{eq-labellike}
	\begin{split}
	&p(y_i=o_m,\exists o_m\in q^t|L_s^t=c_n)
	\\
	&=p(y_i=o_m|\exists o_m\in q^t,L_s^t=c_n)p(\exists o_m \in q^t|L_s^t=c_n)
	\end{split}
\end{equation}
The first term is a detection likelihood while the second term is a tagging likelihood. They can be statistically summarized using the detection results from GroundingDINO and tagging results from RAM.
We follow the equation (\ref{eq-labellike}) to construct a label likelihood matrix over $o_m \in \mathcal{L}_o$ and $c_n \in \mathcal{L}_c$. In the ScanNet training set, we sample $35,000$ image frames with tagging results, detection results, and ground-truth annotation to summarize the statistics.
In the Bayesian update step in equation (\ref{eq-scorelike}), the label likelihood between each pair of $\{o_m,c_n\}$ can be queried from the constructed label likelihood matrix.

\begin{figure}[ht]
	\centering	
	\includegraphics[width=0.9\columnwidth]{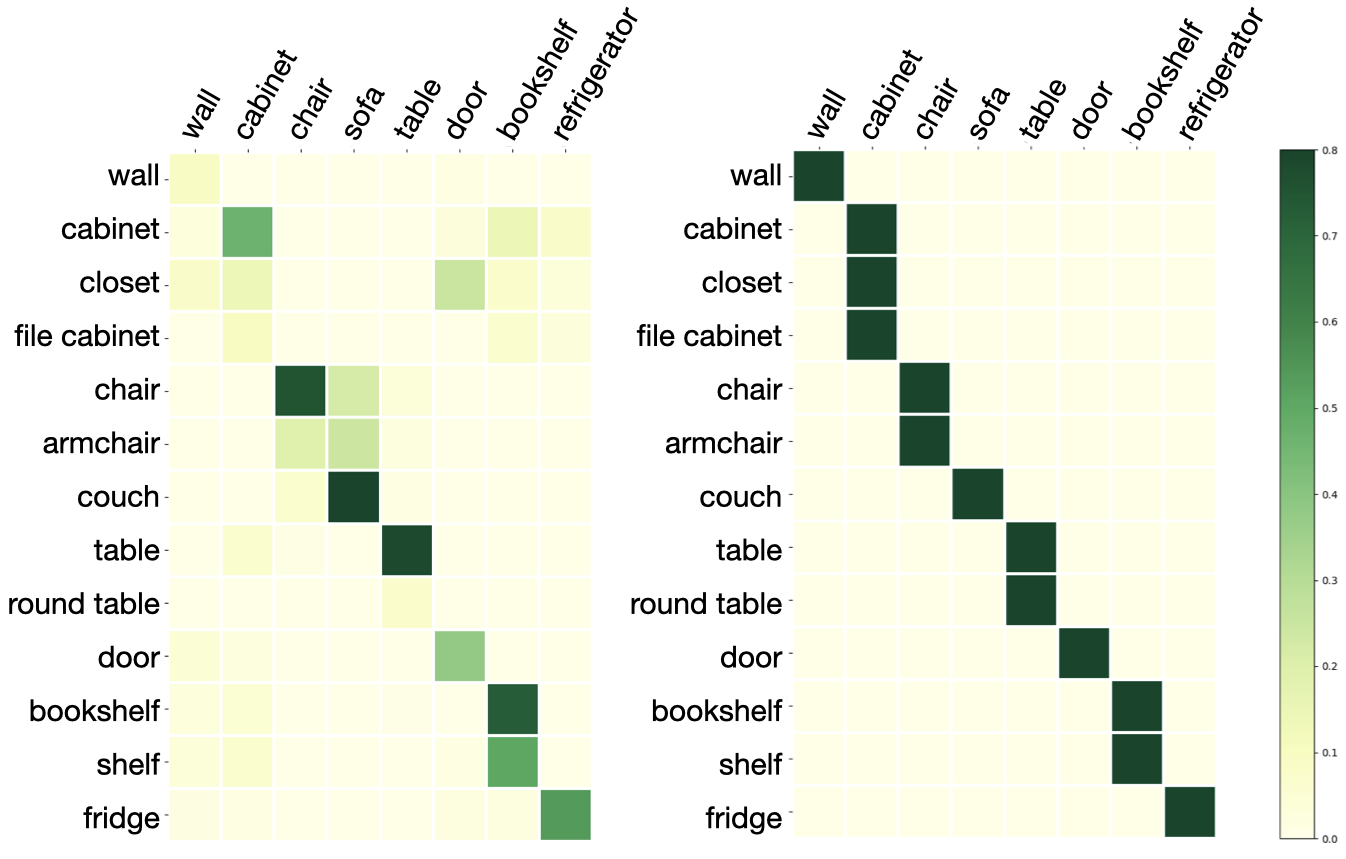}
	\caption{The label likelihood matrix $p(y_i=o_m,\exists o_m\in q^t|L_s=c_n)$ summarized in ScanNet is shown on the left. 
	Each column represents a specific true semantic class $c_n$, while each row represents a measured open-set label $o_m$. On the right, it is a manually assigned likelihood matrix.
	}\label{fig-likelihood}
\end{figure}

As shown in Figure \ref{fig-likelihood}(a), parts of the constructed label likelihood matrix are visualized, while the complete likelihood matrix involves the entire $\mathcal{L}_o$ and $\mathcal{L}_c$.
For comparison, we construct a manually assigned label likelihood matrix similar to Kimera. As shown in Figure \ref{fig-likelihood}, the statistic summarized likelihood matrix is quite different from the manually assigned one. 
In the statistical label likelihood, each semantic class can be detected by its similar open-set labels at various probabilities. Those cells beyond the diagonal can also have likelihood values, indicating the probability of falsely measured labels. The summarized likelihood matrix following equation (\ref{eq-labellike}) describes the probability distribution of label measurements reasonably.

In actual implementation, the multiplicative measurement update in equation (\ref{eq-update}) frequently generates over-confident probability distribution, which is also reported in Fusion++\cite{fusion++2018McCormac}. It causes $p(L_s^t)$ can be easily dominated by the latest measurement $z^t_k$ even if previous label measurements are all different with $z^t_k$. As a result, in the measurement update, we propagate the probability distribution by weighted addition.
\begin{equation}
	p(L_s^t) = \frac{p(z^t_k|L^t_s)+(t-1)\bar{p}(L^t_s)}{t}
\end{equation}

\vspace{-0.3cm}Then, the predicted semantic class for each instance at frame $t$ is $\arg\max_{c_n} p(L_s^t=c_n)$.

\section{Instance refinement}
\begin{figure}[h]
	\includegraphics[width=\columnwidth]{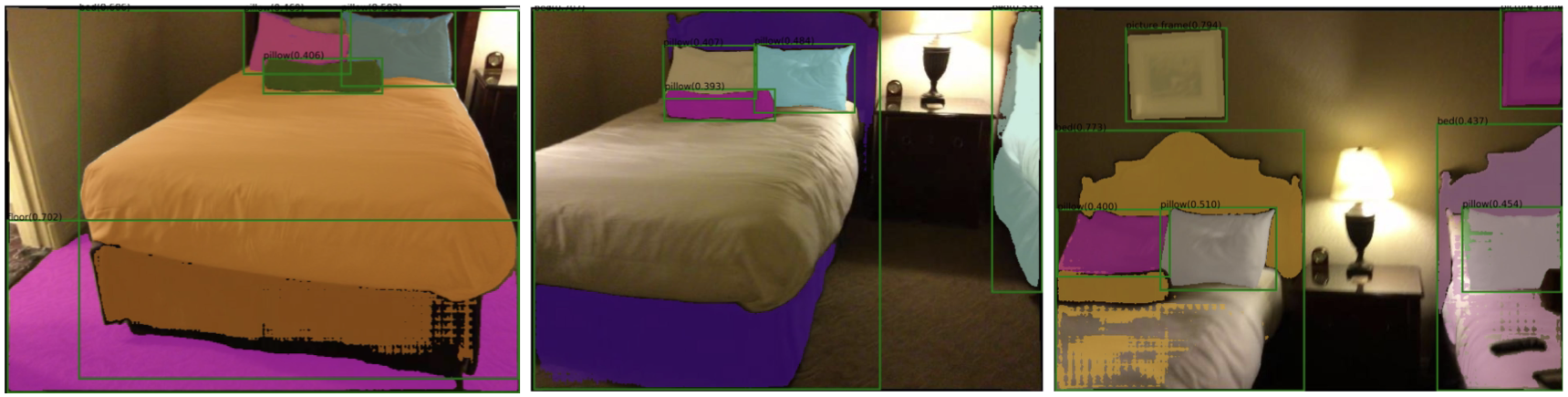}
	\caption{An example of an inconsistent instance mask generated from SAM. In each of the three frames, different areas of the bed are segmented.}
	\label{fig-noisemask}
	\vspace{-0.7cm}
\end{figure}

\subsection{Merge over-segmentation}
Although SAM has demonstrated promising segmentation on a single image, it generates inconsistent instance masks at changed viewpoints, as shown in Figure \ref{fig-noisemask}. The inconsistent masks prevent a correct data association between detections and observed instances. Those mismatched detections are initialized as new instances and cause over-segmentation, as shown in Figure \ref{fig-oversegment}(a).

\newlength{\figwidth}
\setlength{\figwidth}{0.46\columnwidth}
\begin{figure}[ht]
	\centering
	\begin{subfigure}[h]{\figwidth}
		\includegraphics[width=\columnwidth]{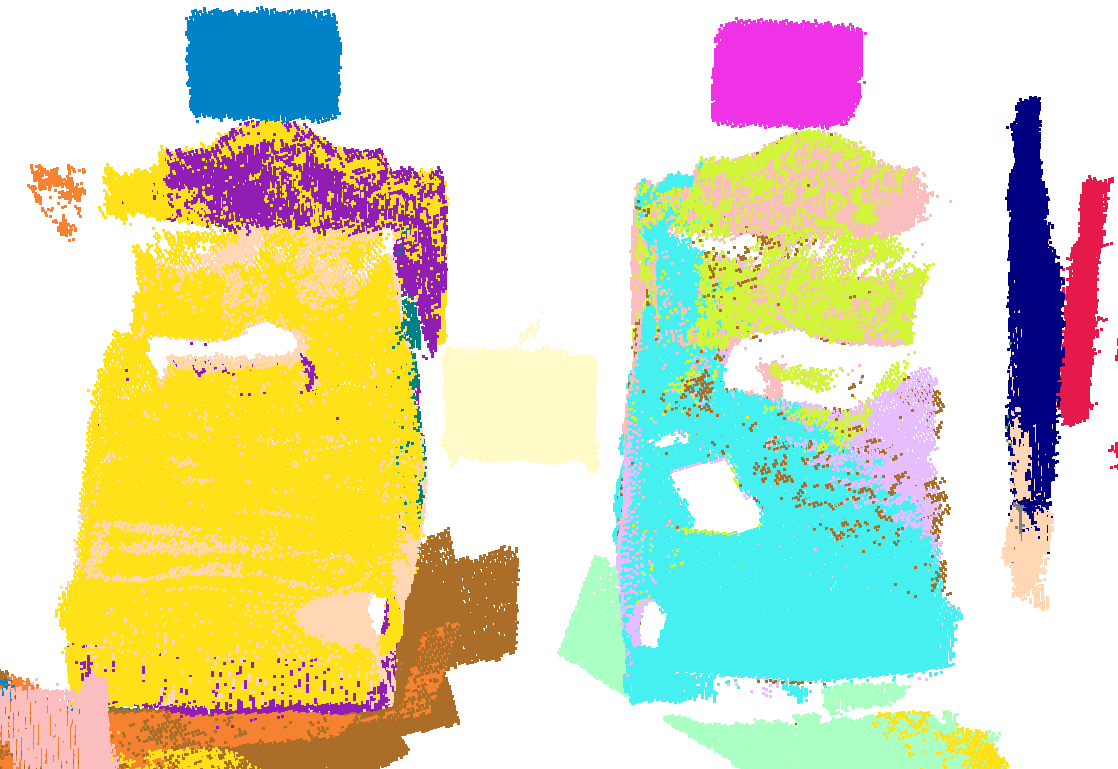}
		\caption{}
	\end{subfigure}
	\begin{subfigure}[h]{\figwidth}
		\includegraphics[width=\columnwidth]{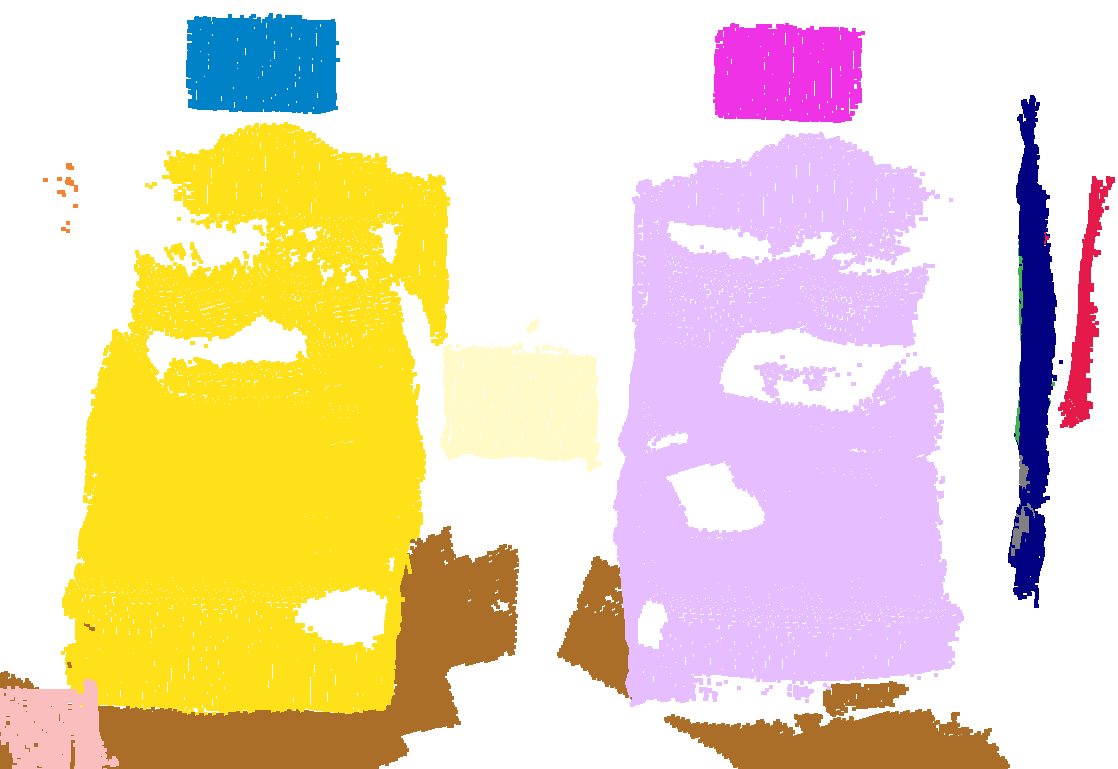}
		\caption{}
	\end{subfigure}
	\caption{The visualization shows instance voxel grid map (a) before and (b) after the merge.}
	\label{fig-oversegment}
\end{figure}

The inconsistent instance mask is a natural limitation for image-based segmentation networks, including SAM and Mask R-CNN. To address it, we utilize spatial overlap information to merge the over-segmentation.
For a pair of instances $\{\mathbf{f}_a,\mathbf{f}_b\}$ at detection frame $t$, where $\mathbf{f}_a$ is volumetric larger than $\mathbf{f}_b$, their semantic similarity $\sigma(\mathbf{f}_a,\mathbf{f}_b)$ and 3D IoU $\Omega(\mathbf{f}_a,\mathbf{f}_b)$ are calculated,
\begin{equation}
	\sigma(\mathbf{f}_a,\mathbf{f}_b)=\bar p(L_s^t(a)) \cdot \bar p(L_s^t(b))
	\label{eq-semanticverify}
\end{equation}
\begin{equation}
	\Omega(\mathbf{f}_a,\mathbf{f}_b) = \frac{\hat{\mathbf{v}}_a\cup \mathbf{v}_b}{\mathbf{v}_b}
\end{equation}
where $\bar p^t(L_s(a))$ is the normalized semantic distribution, ${\mathbf{v}}$ is an instance voxel grid map and $\hat{\mathbf{v}}$ is the inflated voxel grid map. The voxel inflation is designed to enhance 3D IoU for instances with sparse volume. It can be directly generated by scaling the length of each voxel in $\mathbf{v}$.
If the semantic similarity and 3D IoU are both larger than the corresponding thresholds, $\mathbf{f}_b$ is integrated into the voxel grid map of $\mathbf{f}_a$ and further cleared from the instance map. As shown in Figure \ref{fig-oversegment}(b), over-segmented instances caused by inconsistent object masks are merged.

\vspace*{-0.3cm}
\subsection{Instance-geometry fusion}

\setlength{\figwidth}{0.46\columnwidth}
\begin{figure}[h]
	\centering
	\includegraphics[width=0.9\columnwidth]{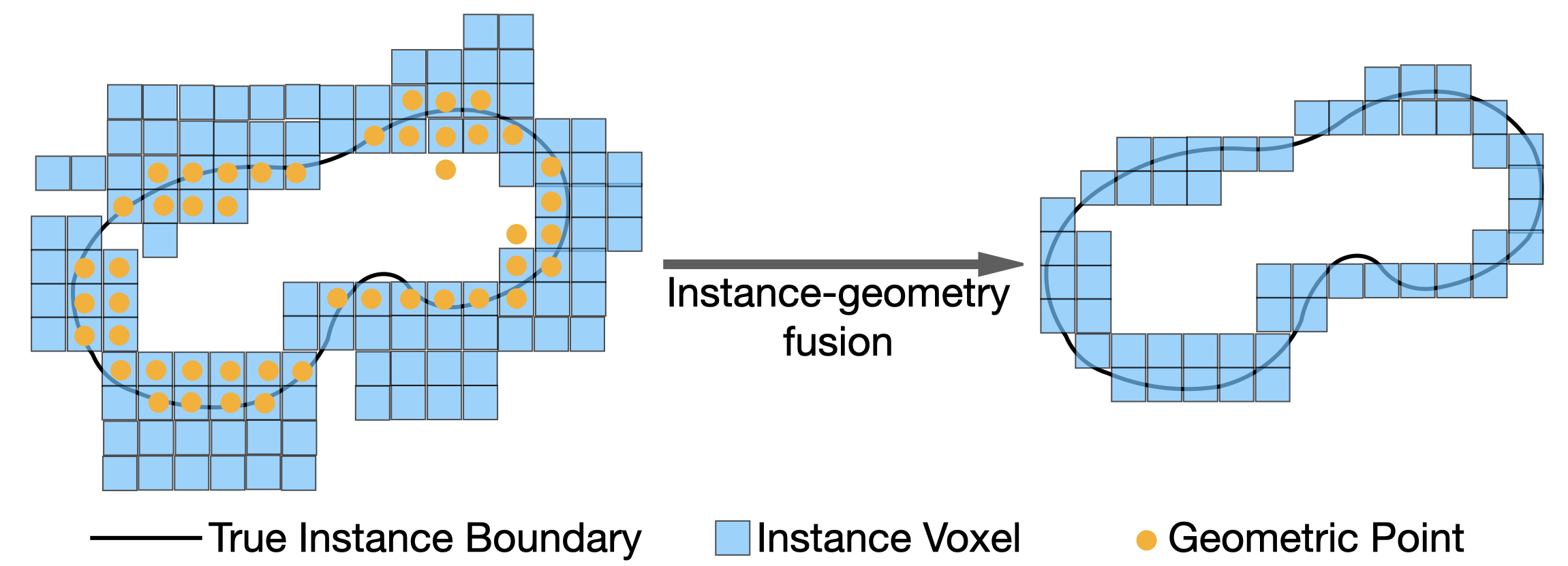}
	\vspace{-0.2cm}
	\caption{Illustration of the instance-geometry fusion. Geometric points are extracted from the global map.} 
	\label{fig-noisyvoxel}
	\vspace{-0.3cm}
\end{figure}

The instance-wise voxel grid map can contain voxel outliers due to noisy depth images being integrated. On the other hand, the global TSDF map is a precise 3D geometric representation. It is because the global TSDF map integrates all the observed volumes in each RGB-D frame, while instance volume only integrates a masked RGB-D frame if the corresponding instance is correctly detected. To filter voxel outliers, we fuse instance-wise voxel grid map $\mathbf{v}$ with the point cloud $\mathbf{P}$ extracted from the global TSDF map. 
As shown in Figure \ref{fig-noisyvoxel}, those voxels in $\mathbf{v}$ that are not occupied by any point in $\mathbf{P}$ are outliers and have been removed. The fused voxel grid map represents the instance volume precisely.

\section{Experiment}

We chose the public dataset ScanNet and SceneNN to evaluate the semantic mapping quality. In ScanNet, 30 scans from its validation set are used. We evaluated its semantic instance segmentation by average precision (AP). In another experiment, we selected 5 scans from SceneNN and evaluated the generalization ability of our method. The SceneNN scans are also used by the previous method \cite{voxblox++2019Grinvald}. In all the experiments, camera poses are provided by the dataset.

We {compared} our method with Kimera
\footnote[2]{https://github.com/MIT-SPARK/Kimera-Semantics} 
and a self-implemented Fusion++. To enable Kimera to read open-set labels $\mathcal{L}_o$, we converted each label in $\mathcal{L}_o$ to a semantic class in NYUv2 label-set $\mathcal{L}_c$. The hard association between $\{\mathcal{L}_o,\mathcal{L}_c\}$ are decided by an academic ChatGPT 
\footnote[3]{https://chatgpt.ust.hk}. 
Then, Kimera can reconstruct a point cloud with semantic labels. To further generate instance-aware point cloud, we employed the geometric segmentation method known as "Cluster-All" \cite{douillard2011segmentation}. It clusters the nearby points with identical semantic labels into an instance. Cluster-All is applied as a post-processing step on the reconstructed semantic map from Kimera. Notice that Cluster-All is very similar to the post-processing module provided by Kimera. But we use Cluster-All for convenient implementation.
Meanwhile, Fusion++ is implemented based on our system modules. Compared with the original Fusion++ method, the main difference is that our implemented version does not maintain a foreground probability for each voxel. Instead, we updated the voxel's weight and filter background voxels using their weights. In experiments with traditional object detection, we used a Mask R-CNN backbone 
with FPN101 image backbone. We evaluated a pre-trained Mask R-CNN and a fine-tuned Mask R-CNN. The pre-trained one is trained in COCO instance segmentation dataset, while we also fine-tuned it using ScanNet dataset.

In implementation, we utilized Open3D\cite{Zhou2018open3d} toolbox to construct the global TSDF map and instance-wise voxel grid map. The global TSDF map is integrated for every RGB-D frame, while our method and all baselines run in every $10$ frames to integrate the detected instances. In all the experiments, the RGB-D images are in $640 \times 480$ dimension and the voxel length is set to be $1.5$ cm.
The experiment is run on an Intel-i7 computer with Nvidia RTX-3090 GPU in an offline fashion.

\begin{table*}[t]
	\centering
	\vspace{+0.5cm}
	\begin{tabular}{c|m{0.3cm} m{0.3cm} m{0.3cm} m{0.3cm} m{0.3cm} m{0.3cm} m{0.3cm} m{0.3cm} m{0.3cm} m{0.3cm} m{0.3cm} m{0.3cm} m{0.3cm} m{0.3cm} m{0.3cm} m{0.3cm} m{0.3cm} m{0.3cm}|c c }
	 \hline
	  Method & cab. & bed & cha. &sof. & tab. & door & win. & bkf. &pic. & cou. & desk &cur. & ref. &show. & toi.& sink & bath. & oth.& $\text{mAP}_{50}$\\
	  \hline 
	  M-CNN\& Kimera & 0.0 & 6.4 &10.0 & 25.1 & 17.3 & 0.0 &0.0 &0.0  &0.0 & 0.0 & 0.0 &0.0 & 24.6 & 0.0 & 10.4 &4.3 &0.0 &0.0 & 5.4\\
	  M-CNN\& Fusion++ &0.0 & 27.1 & 3.7 & 14.7 & 4.4 &0.0 & 0.0 &0.0 &0.0 &0.0 &0.0 & 0.0 & 23.9 & 0.0 & 46.6 & 20.0 & 0.0 & 0.0 & 7.8\\
	  \hline
	  M-CNN$^\ast$\& Kimera & 27.8 & \textbf{55.5} &18.7 & 0.0 & 0.0 & 16.5 &\textbf{33.1} &31.2  &12.9 & 23.3 & 3.9 &\textbf{26.0} & 0.0 & 75.0 & 60.0 &22.7 &60.0 &0.0 & 25.9\\
	  M-CNN$^\ast$\& Fusion++ &7.1 & 22.0 & 31.2 & 0.0 & 13.3 &12.5 & 15.0 &11.5 &28.5 &0.0 &0.0 & 18.1 & 0.0 & 0.0 & 0.0 & 40.1 & 0.0 & 0.0 & 11.1\\
	  \hline
	  G-SAM \&Kimera & \textbf{32.5} & 21.0 &16.8 & \textbf{54.5} & 21.7& \textbf{26.0} & \textbf{31.5} &45.3  &21.9 & \textbf{8.7} & {3.9} & \textbf{24.8} & 29.6 & 0.0 & 50.0 &{9.4} &46.2 &0.0 & 24.7\\
	  G-SAM \& Ours& 4.6 & {46.2} &\textbf{49.2} &{39.6} &\textbf{37.3} &{19.5} &{12.0} & \textbf{50.4} & \textbf{44.0} &{3.8} &\textbf{8.9} &{15.5} &\textbf{66.7} &\textbf{82.2} &\textbf{100.0} &\textbf{41.5} & \textbf{75.0} & \textbf{30.7} & \textbf{40.3}\\
	\hline
	\end{tabular}
	 \caption{Evaluate Kimera, the implemented Fusion++, and the proposed method on ScanNet using 30 validation scans. We report $\text{AP}_{50}$ at $50\%$ IoU threshold for each semantic class. M-CNN denotes Mask R-CNN, M-CNN$^\ast$ is fine-tuned Mask R-CNN, while G-SAM refers to RAM-Grounded-SAM.
	 }
	 \label{tab:eval}
	 \vspace*{-0.3cm}
	\end{table*}


\setlength{\figwidth}{0.47\columnwidth}
\begin{figure*}[t]
	\centering

	\begin{subfigure}[h]{\figwidth}
		\includegraphics[width=\columnwidth]{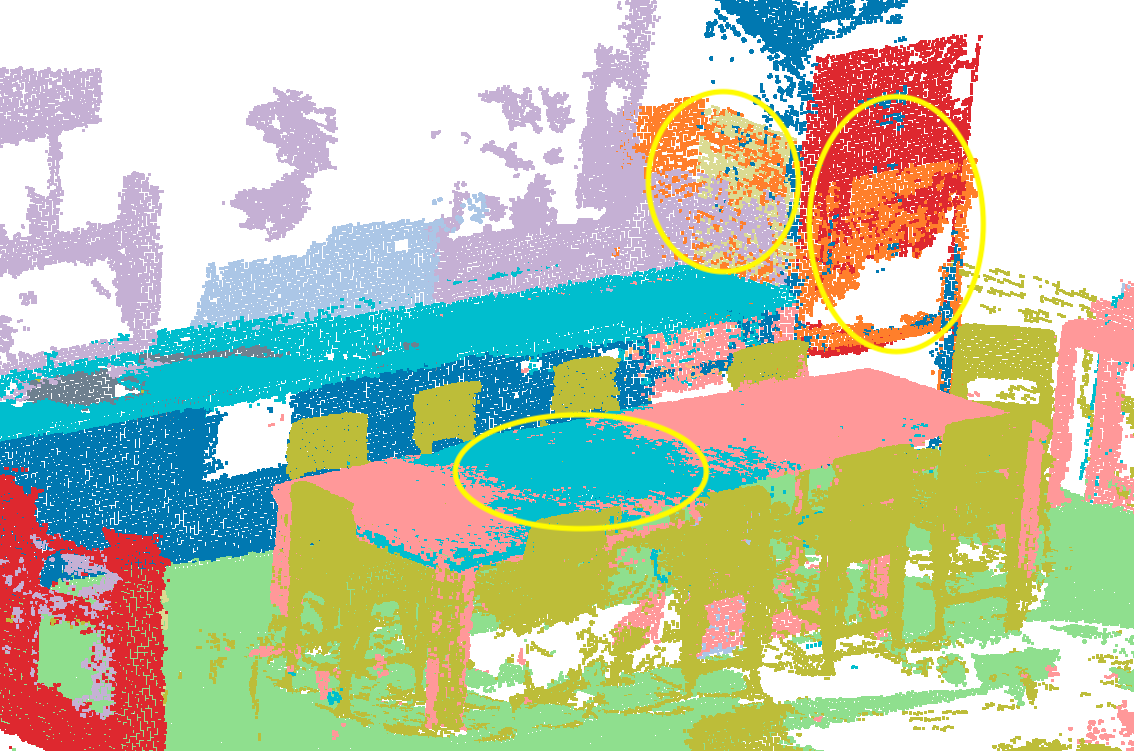}
	\end{subfigure}	
	\begin{subfigure}[h]{\figwidth}
		\includegraphics[width=\columnwidth]{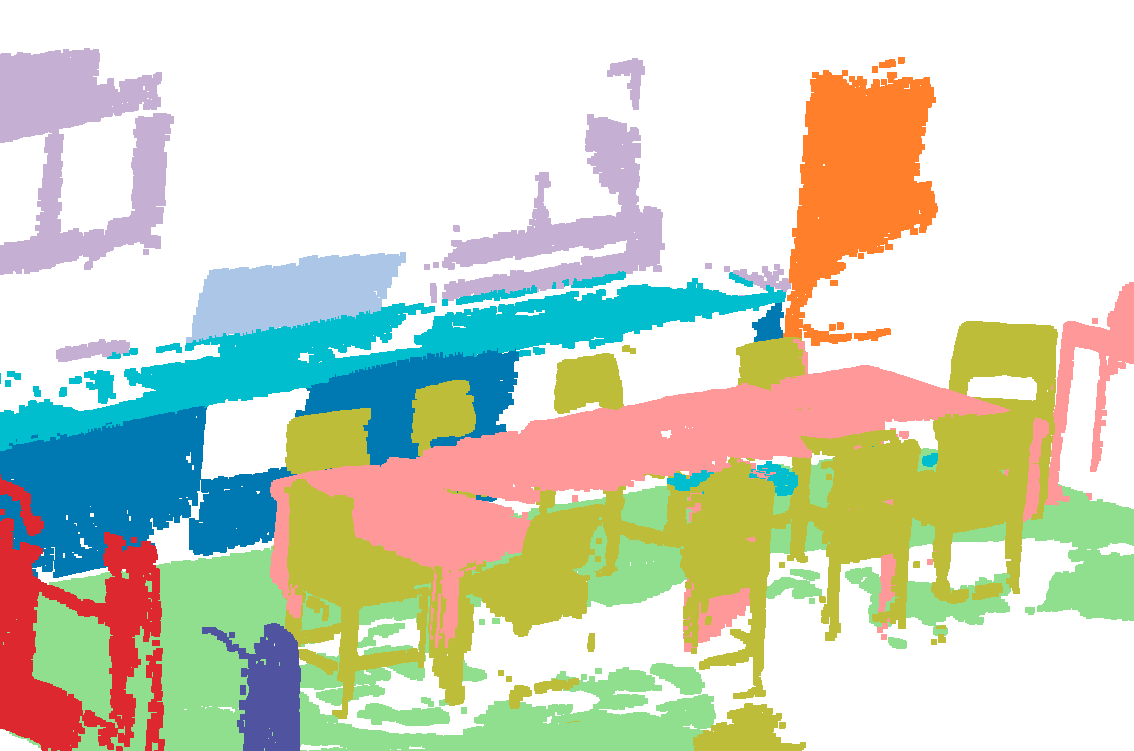}
	\end{subfigure}
	\begin{subfigure}[h]{\figwidth}
		\includegraphics[width=\columnwidth]{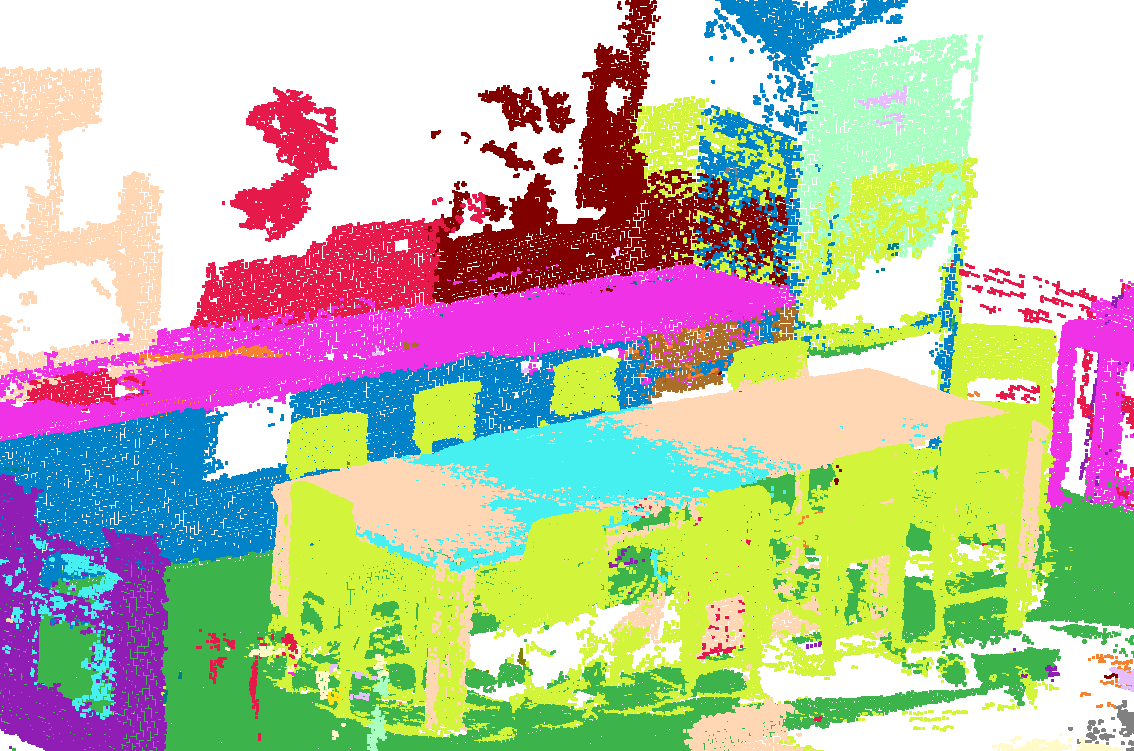}
	\end{subfigure}
	\begin{subfigure}[h]{\figwidth}
		\includegraphics[width=\columnwidth]{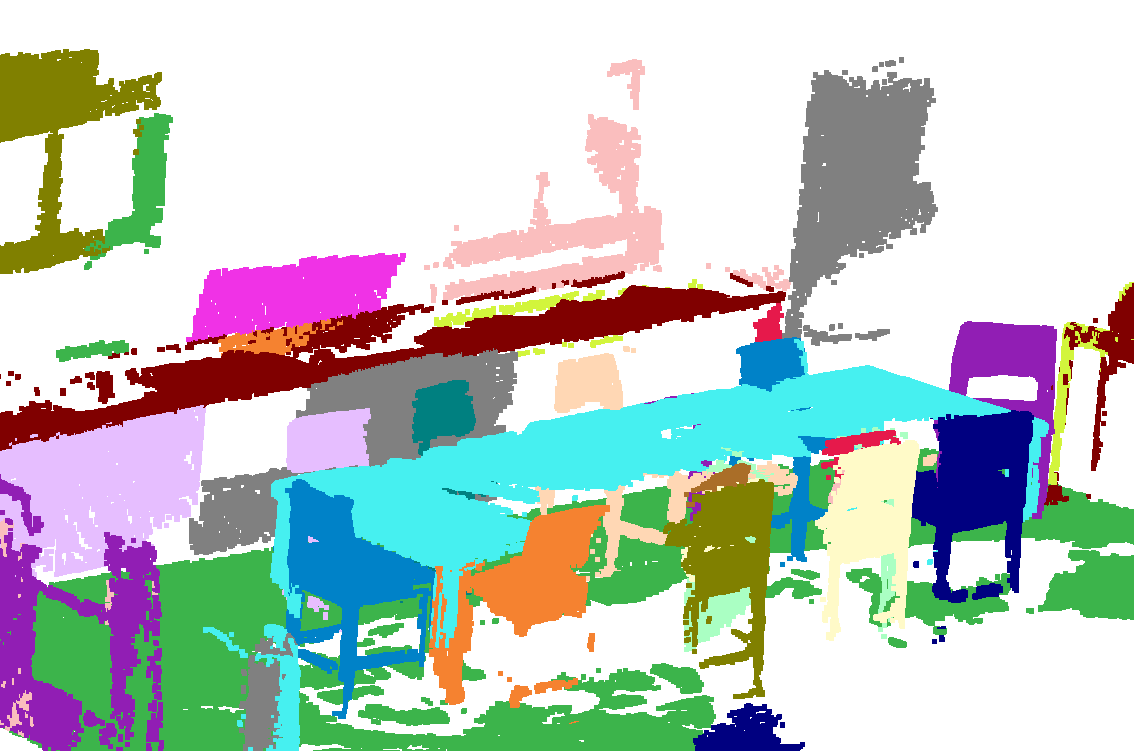}
	\end{subfigure}
	\vspace{+0.1cm}

	\begin{subfigure}[h]{\figwidth}
		\includegraphics[width=\columnwidth]{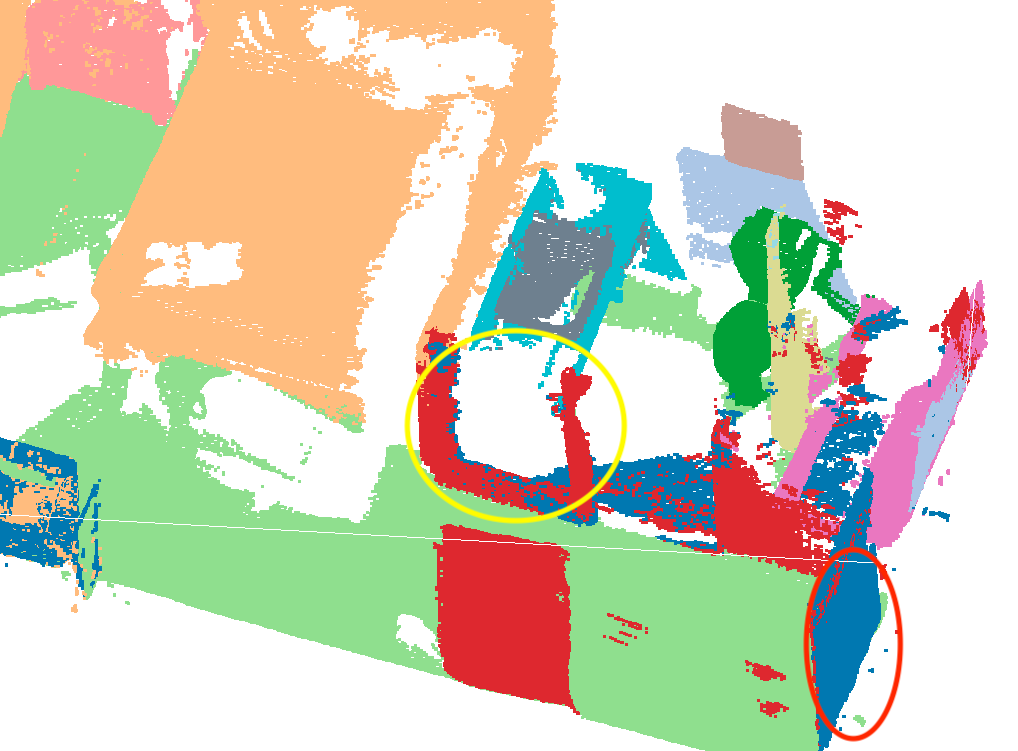}
	\end{subfigure}	
	\begin{subfigure}[h]{\figwidth}
		\includegraphics[width=\columnwidth]{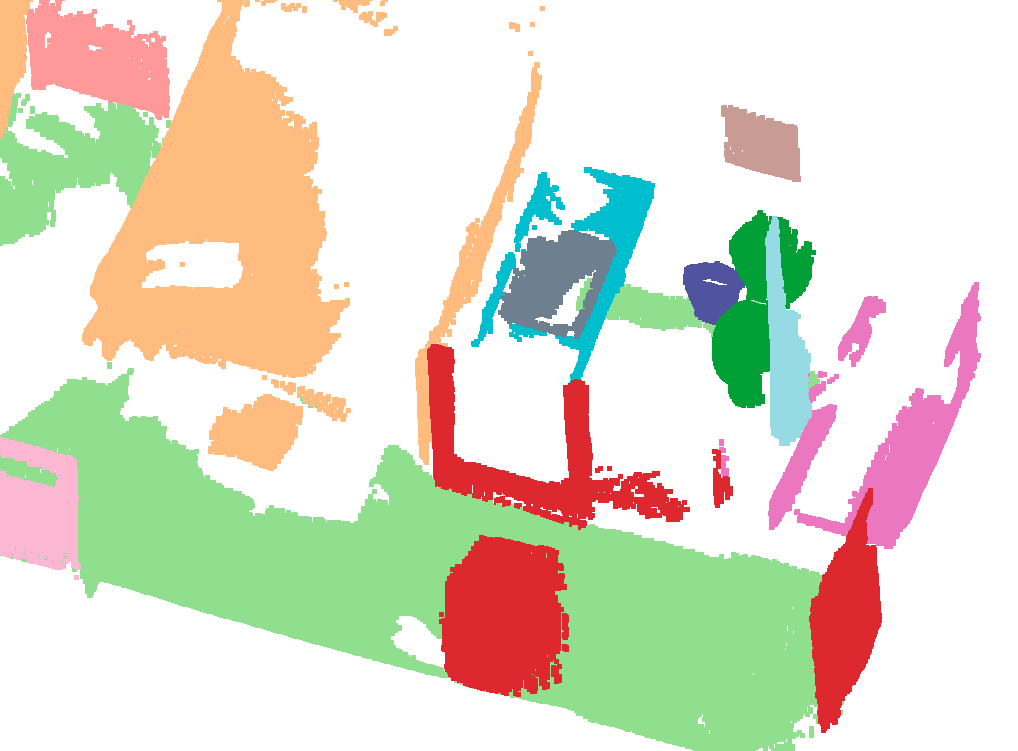}
	\end{subfigure}
	\begin{subfigure}[h]{\figwidth}
		\includegraphics[width=\columnwidth]{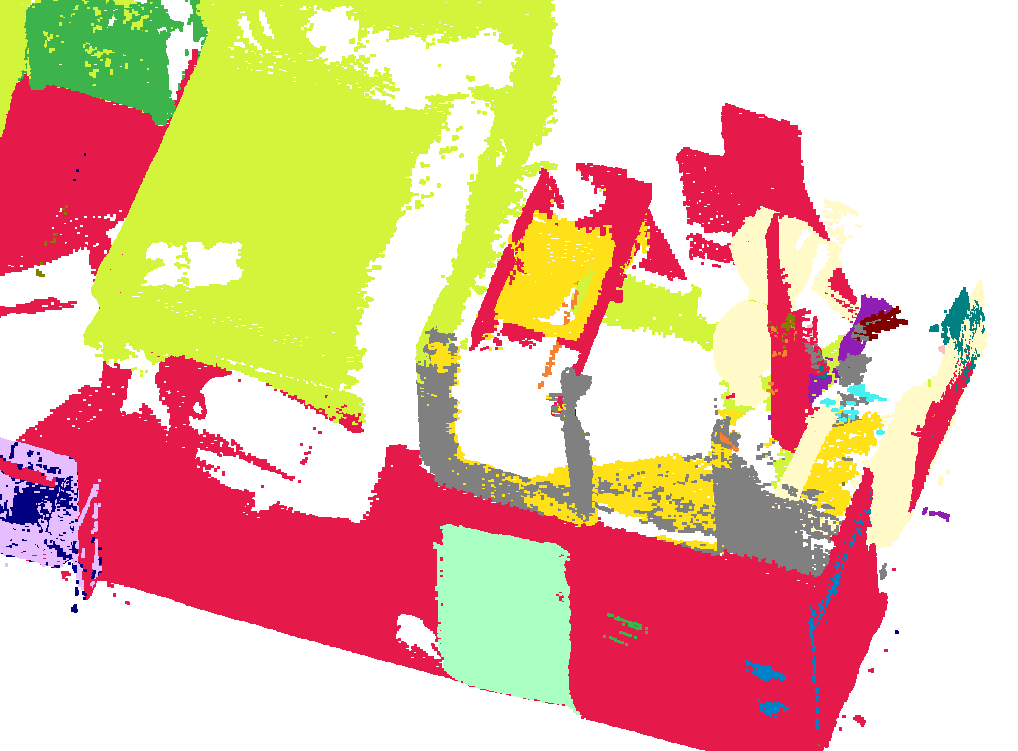}
	\end{subfigure}
	\begin{subfigure}[h]{\figwidth}
		\includegraphics[width=\columnwidth]{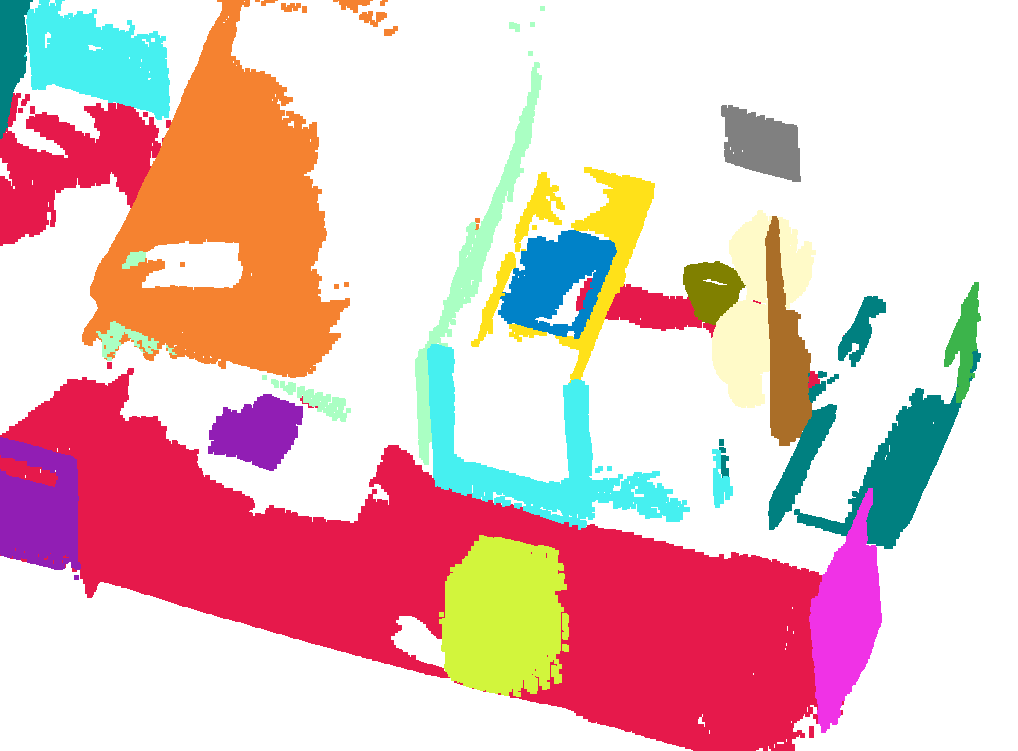}
	\end{subfigure}
	\vspace{+0.1cm}

	\begin{subfigure}[h]{\figwidth}
		\includegraphics[width=\columnwidth]{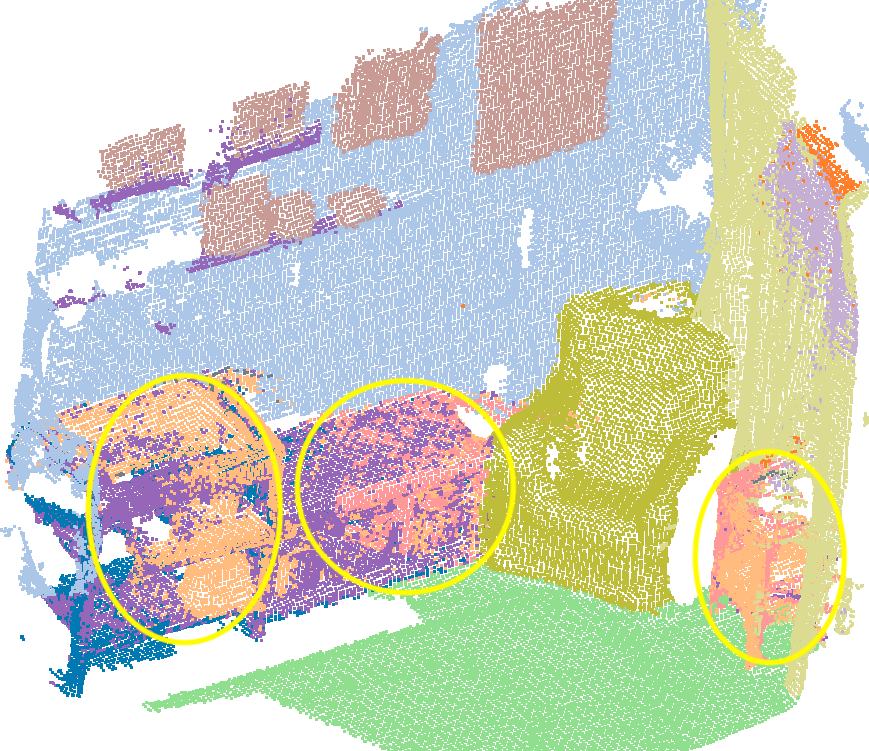}
		\caption{Kimera Semantic}
	\end{subfigure}
	\begin{subfigure}[h]{\figwidth}
		\includegraphics[width=\columnwidth]{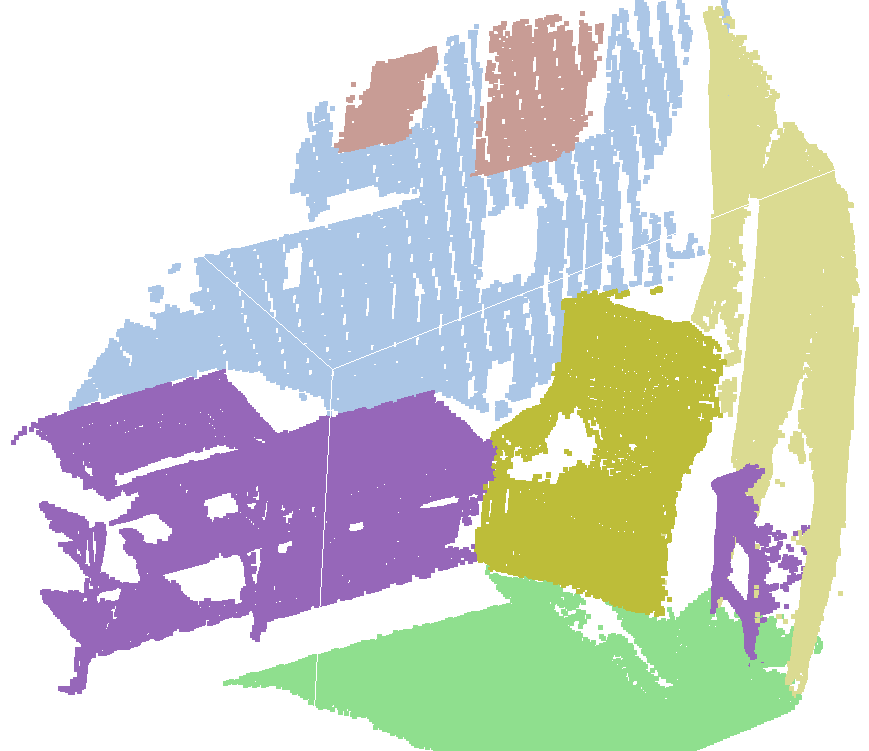}
		\caption{Our Semantic}
	\end{subfigure}
	\begin{subfigure}[h]{\figwidth}
		\includegraphics[width=\columnwidth]{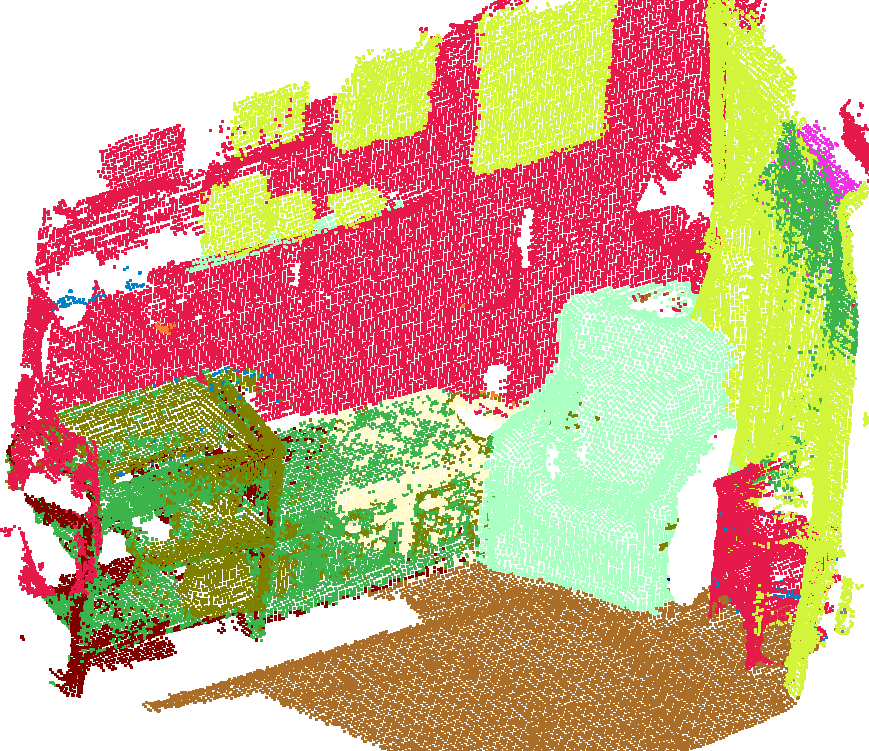}
		\caption{Kimera Instances}
	\end{subfigure}
	\begin{subfigure}[h]{\figwidth}
		\includegraphics[width=\columnwidth]{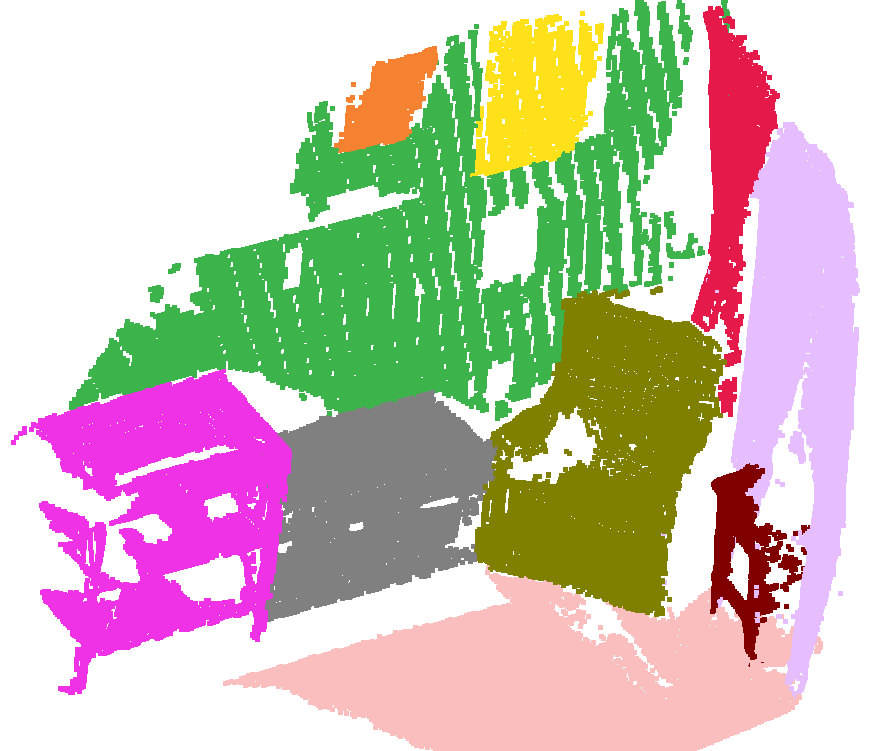}		
		\caption{Our Instances}
	\end{subfigure}
	
	\vspace{-0.2cm}
	\caption{The reconstructed instance map using RAM-Grounded-SAM in ScanNet \emph{scene0011}, \emph{scene0435} and \emph{scene0633} (from top to bottom). The falsely predicted semantic classes in (a) and (b) are highlighted in red circles, while spatial conflicted semantics are in yellow. All semantic maps are colored following the NYUv2 color map and instances are colored randomly. \label{fig:scannet}}
	\vspace{-0.5cm}
\end{figure*}

\vspace{-0.3cm}
\subsection{ScanNet Evaluation}
\vspace{-0.2cm}
In the instance segmentation benchmark, as shown in Table \ref{tab:eval}, semantic mapping based on Mask R-CNN can only reconstruct a few of the semantic categories. It is because the pre-trained Mask R-CNN is trained using COCO label-set and those new semantic classes in NYUv2 label-set are predicted with $0$ AP. Even for those predictable semantic classes, the pre-trained Mask R-CNN suffers from the issue of generalization and achieve low $\text{AP}_{50}$ scores. 
In experiment with fine-tune Mask R-CNN, although the mean AP is improved, they still reconstruct a few of semantic classes with $0$ AP. We also notice that Kimera performs significantly better than the implemented Fusion++. We believe the difference comes from their different map management methods. Unlike Kimera ignores the instance-wise segmentation, Fusion++ maintains instance-wise volumes and requires data association. But the fine-tuned Mask R-CNN still generate detections with noisy instance masks, causing a large amount of false data association. As a result, Fusion++ generates instances with too many over-segmentation and maintains a low AP score.

The results demonstrated that semantic mapping based on supervised object detection can be easily affected by image distribution, label-set distribution and annotation quality.
On the other hand, boosted by pre-trained foundational models RAM-Grounded-SAM, both Kimera and our method reconstructed semantic instances in higher quality than semantic mapping methods based on the supervised object detection.

However, simply replacing object detectors with foundation models could not utilize the maximum potential of the foundation models. Compared with Kimera using RAM-Grounded-SAM, our method achieved $+15.6$ $\text{mAP}_{50}$. The boosted performance comes from two aspects. Firstly, our probabilistic label fusion predicts semantic class in higher accuracy. As shown in Figure \ref{fig:scannet}(a), Kimera predicts semantics of some sub-volumes falsely. Since Kimera updates the label measurements with a manually assigned likelihood probability and ignores the similarity score provided by GroundingDINO, it is easier to be affected by false label measurements.
Secondly, Kimera ignores the instance-level segmentation and reconstructs many over-segmented instances. Some of them are predicted with different semantic labels, as shown in Figure \ref{fig:scannet}(a) and \ref{fig:scannet}(c). However, our method is instance-aware. Each instance volume is maintained separately. Our instance refinement module merges over-segmented instances caused by inconsistent instance masks at changed viewpoints. We further fused instance volume with a global volumetric map. Hence, our instances volumes are spatially consistent and relative precise, as shown in Figure \ref{fig:scannet}(d).

\begin{table}[h]
	\centering
	\begin{tabular}{c| c c c c}
		\hline
		Method & Prm. Aug.& Likelihood & Refine & $\text{mAP}_{50}$\\ 
		\hline
		A & \checkmark & Manual Assign & \checkmark & 35.9\\ 
		B & $\times$ & statistic & \checkmark & 34.1\\		
		C & \checkmark & statistic & $\times$ & 23.4 \\ 
		\textbf{Ours} & \checkmark & statistic & \checkmark & \textbf{40.3} \\
		\hline
	\end{tabular}
	\caption{ Ablation study of FM-Fusion. Prm. Aug. denotes text prompt augmentation.}
	\label{table-ablate}
	\vspace{-0.3cm}
\end{table}

\begin{figure*}[ht]
	\centering
	\vspace{+0.2cm}
	\begin{subfigure}[h]{\figwidth}
		\includegraphics[width=\columnwidth]{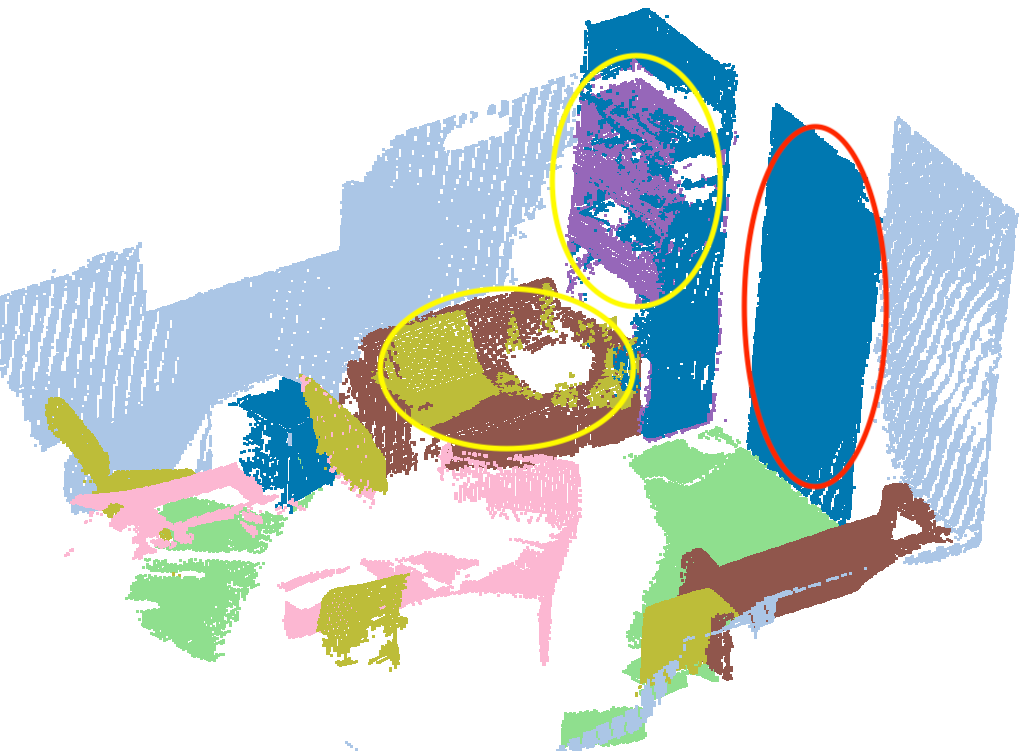}
		\caption{Ablation-A Semantic}
	\end{subfigure}
	\begin{subfigure}[h]{\figwidth}
		\includegraphics[width=\columnwidth]{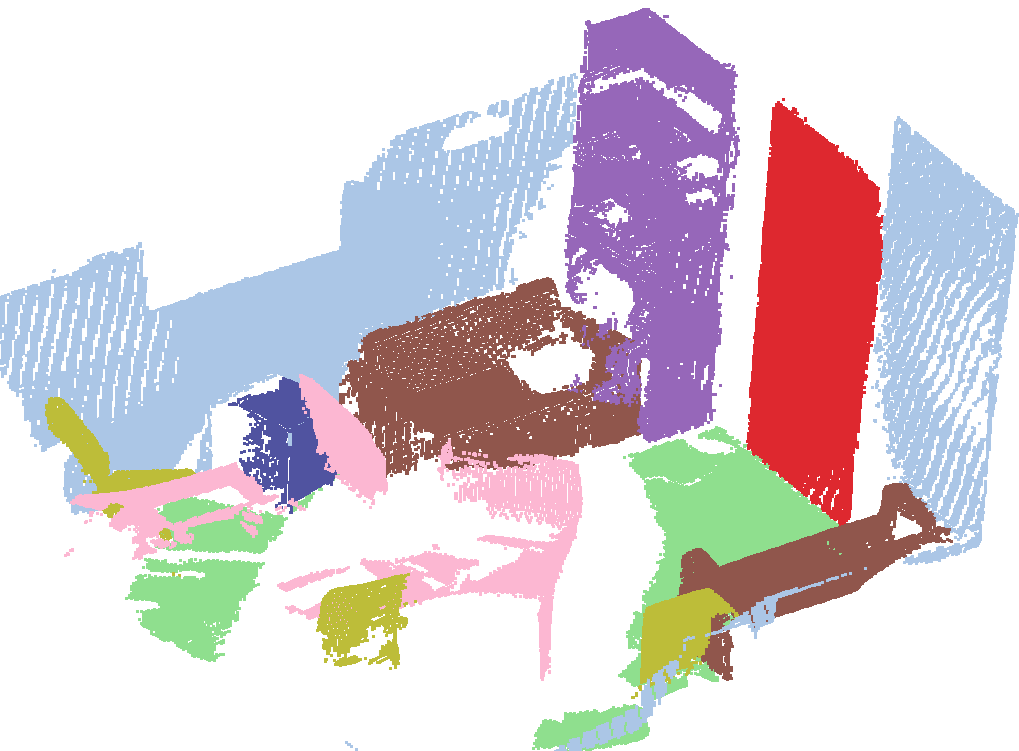}
		\caption{Our Semantic}
	\end{subfigure}
	\begin{subfigure}[h]{\figwidth}
		\includegraphics[width=\columnwidth]{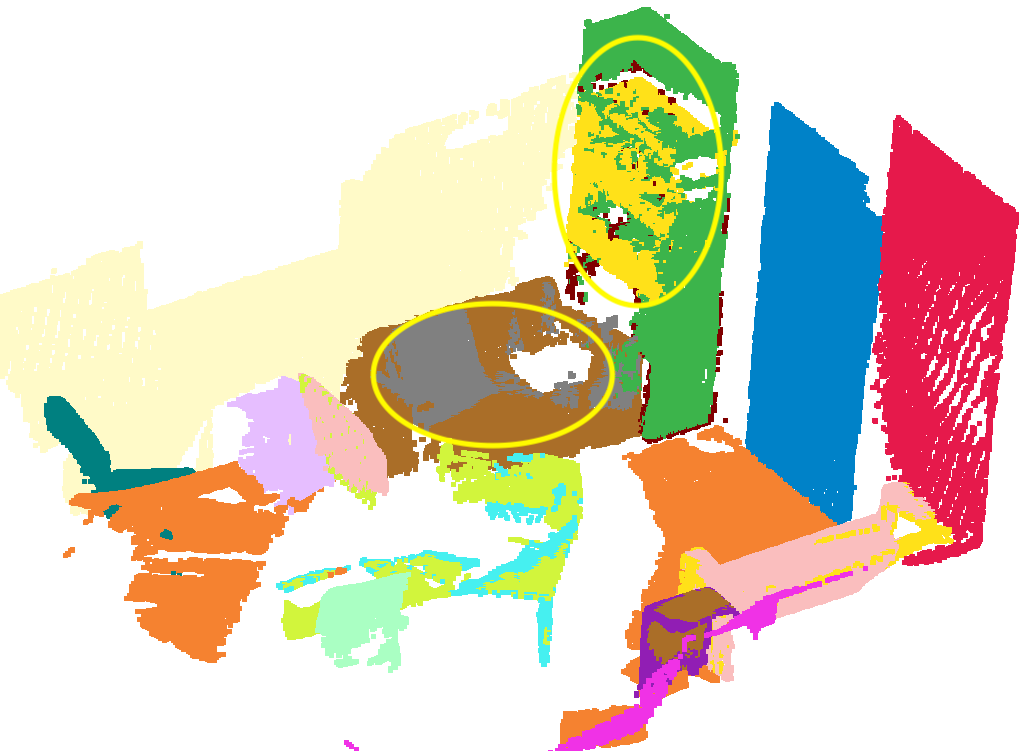}
		\caption{Ablation-A Instances}
	\end{subfigure}
	\begin{subfigure}[h]{\figwidth}
		\includegraphics[width=\columnwidth]{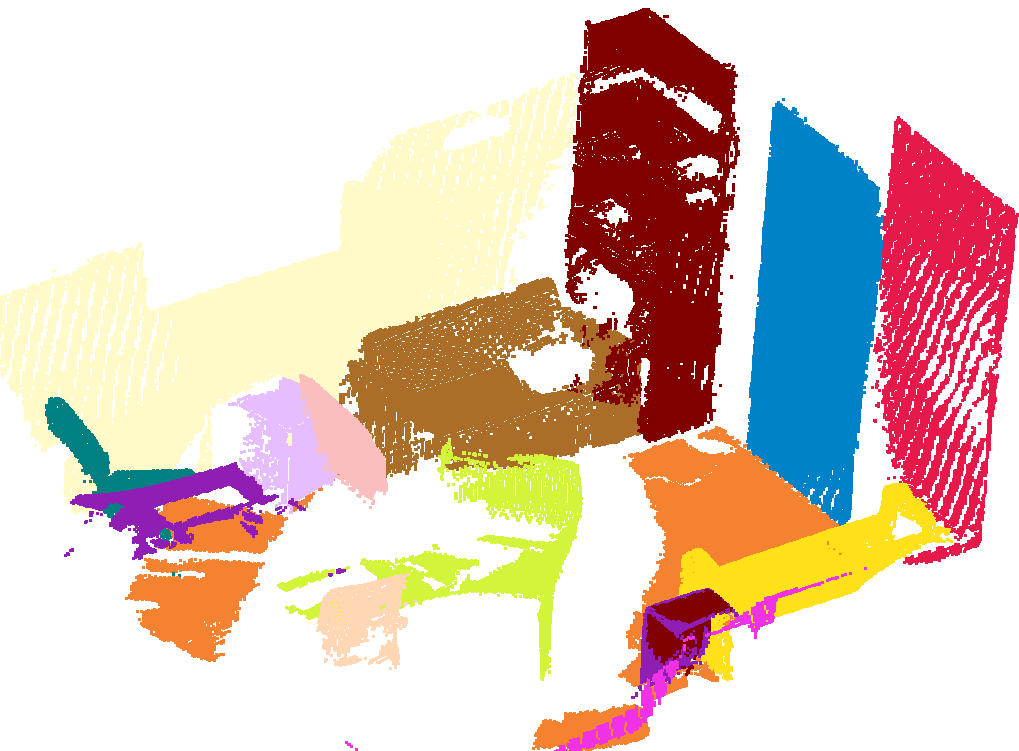}		
		\caption{Our Instances}
	\end{subfigure}
	\caption{ A visualized example at ScanNet \emph{scene0025\_01}. (a) A falsely predicted instance caused by manually assigned likelihood is highlighted in a red circle, while the spatial conflicted semantic predictions are highlighted in yellow. (b) Proposed semantic result. (c) Over-segmentation is highlighted in yellow. (d) Our refined instances. \label{fig:ablate}}
	 \vspace{-0.3cm}
\end{figure*}

The rest of the ScanNet experiment focus on evaluating each module of our method through an ablation study.
As shown in Table.\ref{table-ablate}, the text prompt augmentation, probabilistic label fusion with statistic summarized likelihood, and instance refinement all improve the reconstructed semantic instances.

A visualized example of the Ablation-A is shown in Figure \ref{fig:ablate}. As shown in Figure \ref{fig:ablate}(a), Ablation-A predicts an instance falsely, similar to Kimera. It also predicts overlapped instances with over-confident semantic probability distributions. They can not be merged during refinement due to their low semantic similarity. So, the over-segmented instances can not be merged, as shown in Figure \ref{fig:ablate}(c). On the other hand, our method predicts the corresponding semantic classes correctly. The over-segmented instances are predicted with a similar semantic probability distribution and have been merged successfully, as shown in Figure \ref{fig:ablate}(b) and \ref{fig:ablate}(d). 

\begin{figure}[h]
	\centering
	\includegraphics[width=\columnwidth]{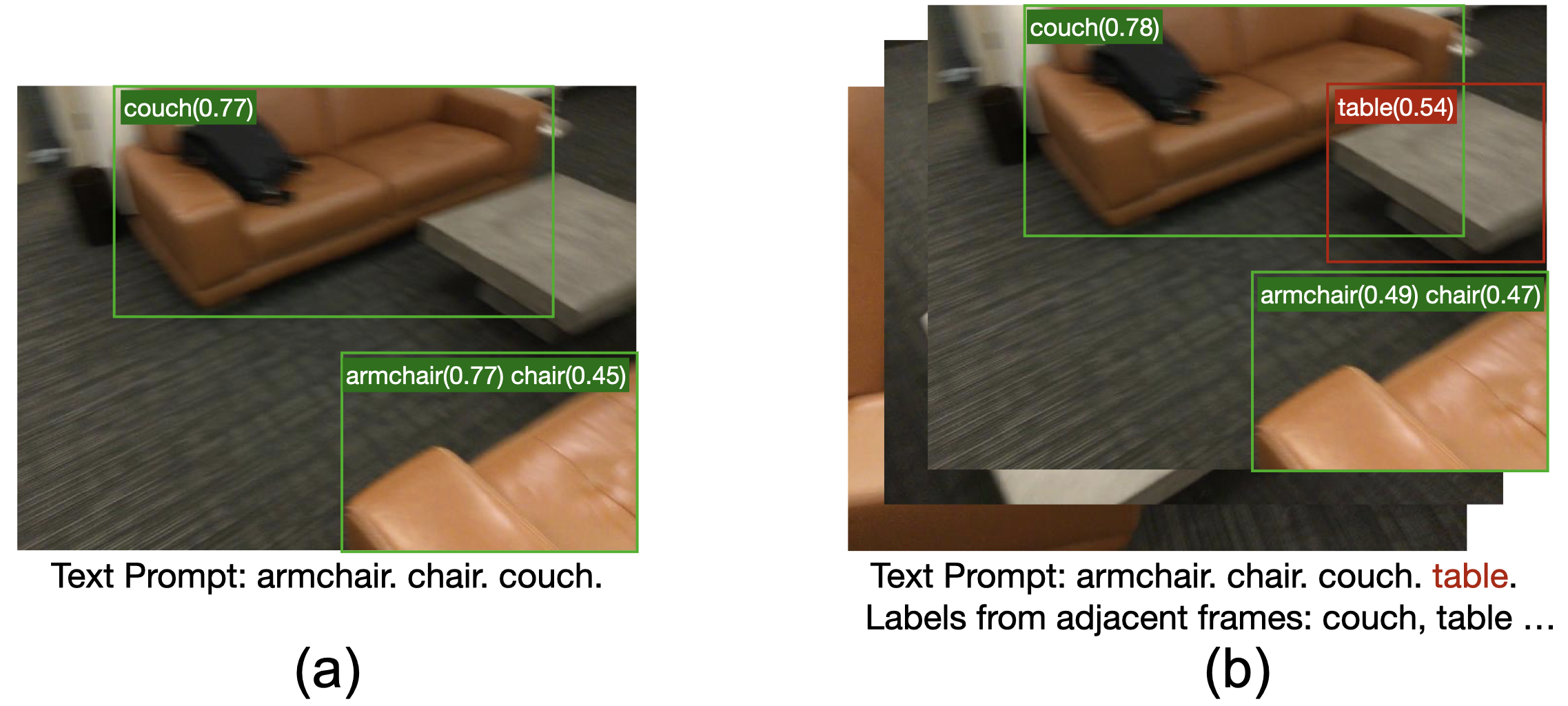}
	\caption{An image of object detection from Ablation-B and our method are shown in (a) and (b). The labels incorporated by text prompt augmentation are highlighted in red. The images are from ScanNet \textit{scene0329}.
	}
	\label{fig-aug} 
	\vspace{-0.3cm}
\end{figure}

As shown in Figure \ref{fig-aug}(a), RAM fails to recognize a table due to the extreme viewpoint, and GroundingDINO cannot detect it either. On the other hand, as illustrated in section \ref{sec-preparedetector}, our method maintains a series of labels $U^t$ that has been detected in previous $5$ frames. If a label in $U^t$ is not given in RAM tags, the corresponding label is added to the text prompt. As shown in Figure \ref{fig-aug}(b), our method detects the table correctly. Beyond miss detecting objects, the incomplete tags from RAM cause false label measurements in other frames. Hence, Ablation-B reconstructs a few instances with a false semantic class. More results can be found in our supplementary video. 

To sum up, simply replacing traditional object detectors with RAM-Grounded-SAM to construct the semantic map improves the semantic mapping performance significantly. However, false label measurements, inconsistent instance masks, and missed tags in the text prompt still exist in foundation models. They limit the performance of semantic mapping. We consider those limitations of foundation models. Compared with Kimera using RAM-Grounded-SAM, our method further improves $\text{mAP}_{50}$ by $+15.6$.

\vspace{-0.3cm}
\subsection{SceneNN evaluation}

\setlength{\figwidth}{0.5\columnwidth}
\begin{figure*}[ht]
	\centering
	\vspace{+0.2cm}
	\begin{subfigure}[h]{\figwidth}
		\includegraphics[width=\columnwidth]{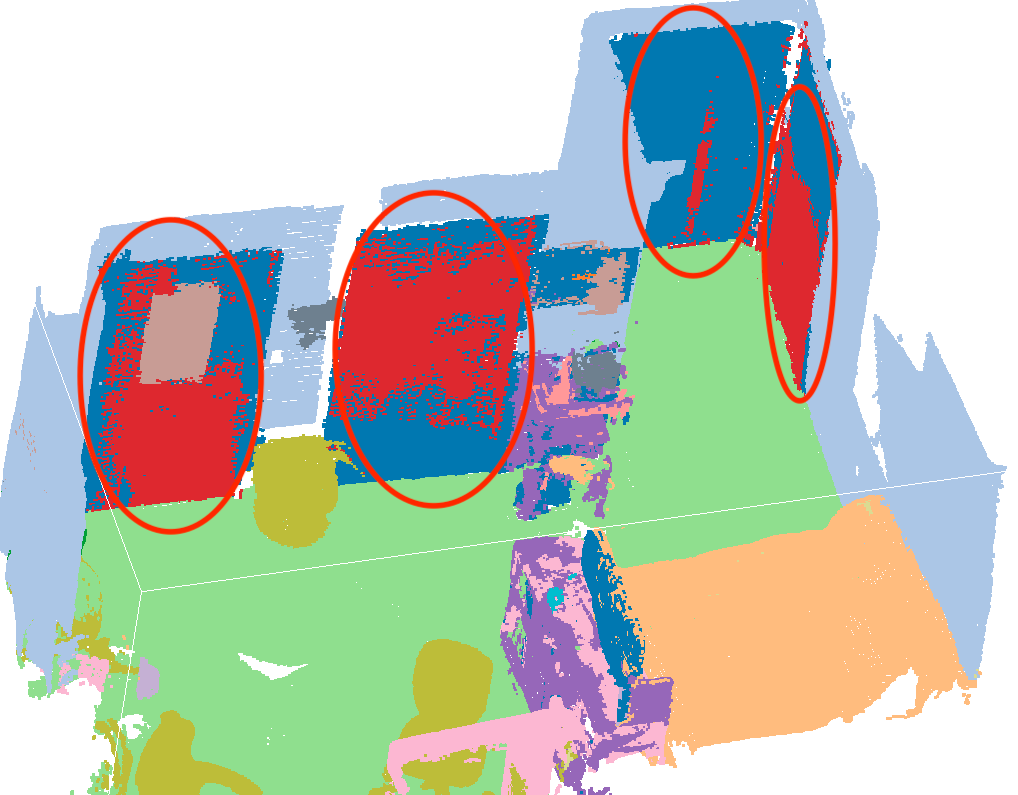}
		\caption{Kimera Semantic}
	\end{subfigure}
	\begin{subfigure}[h]{\figwidth}
		\includegraphics[width=\columnwidth]{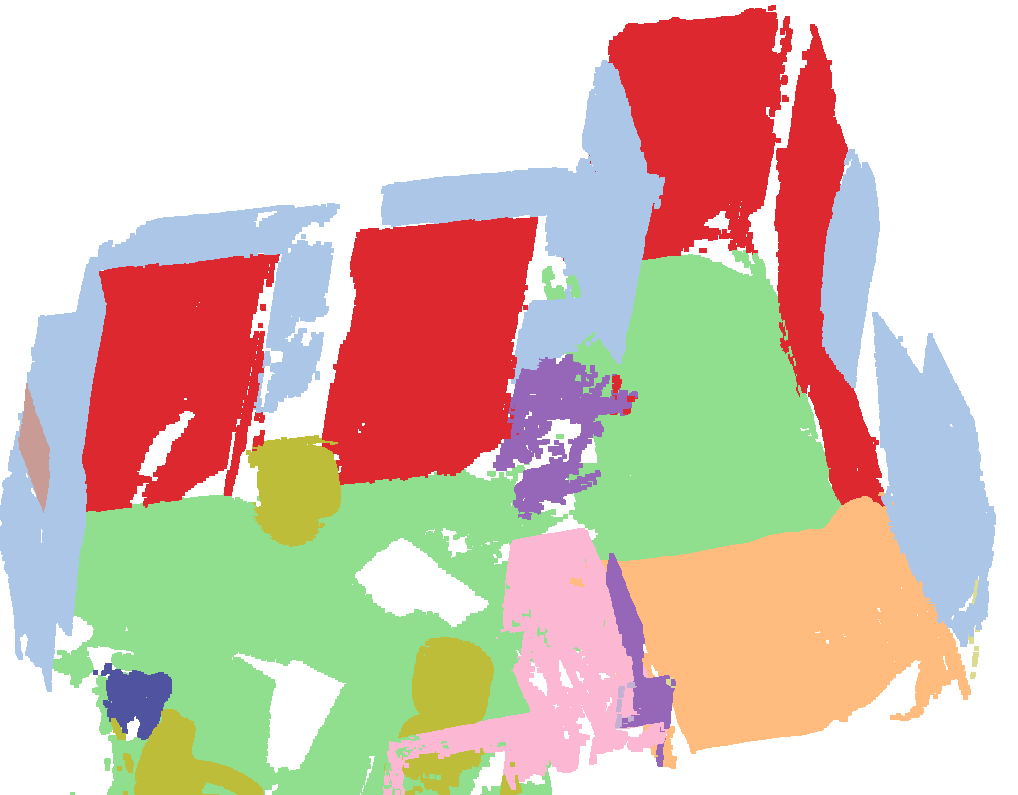}
		\caption{Our Semantic}
	\end{subfigure}
	\begin{subfigure}[h]{\figwidth}
		\includegraphics[width=\columnwidth]{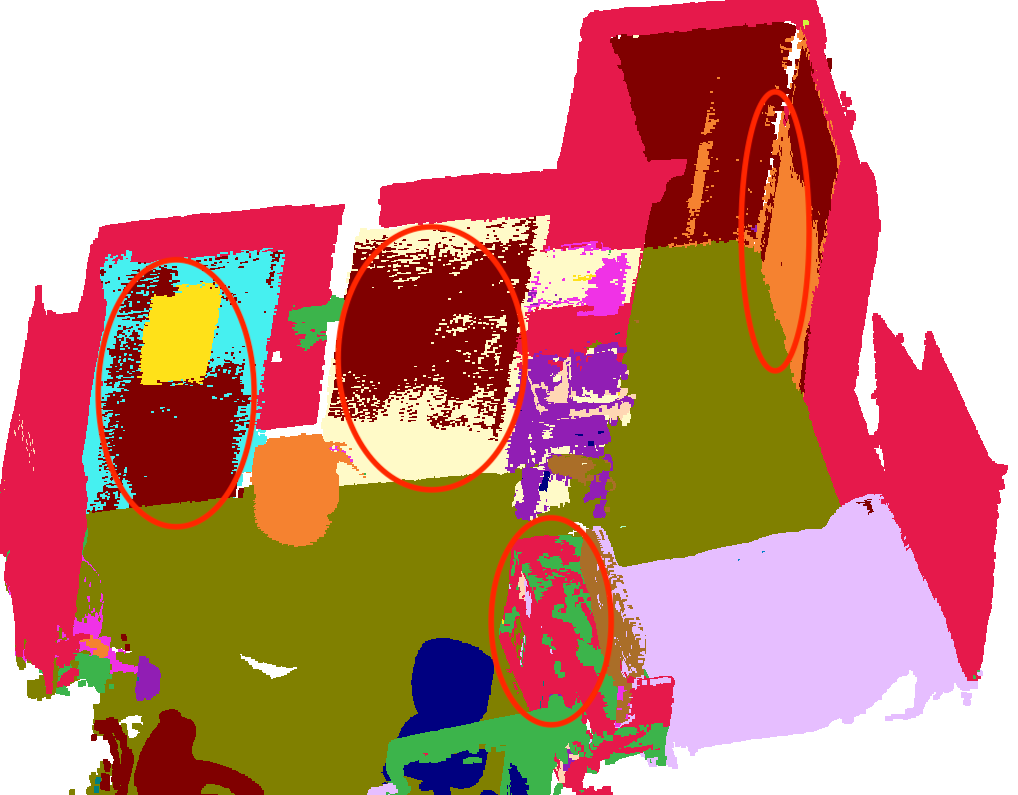}
		\caption{Kimera Instances}
	\end{subfigure}
	\begin{subfigure}[h]{\figwidth}
		\includegraphics[width=\columnwidth]{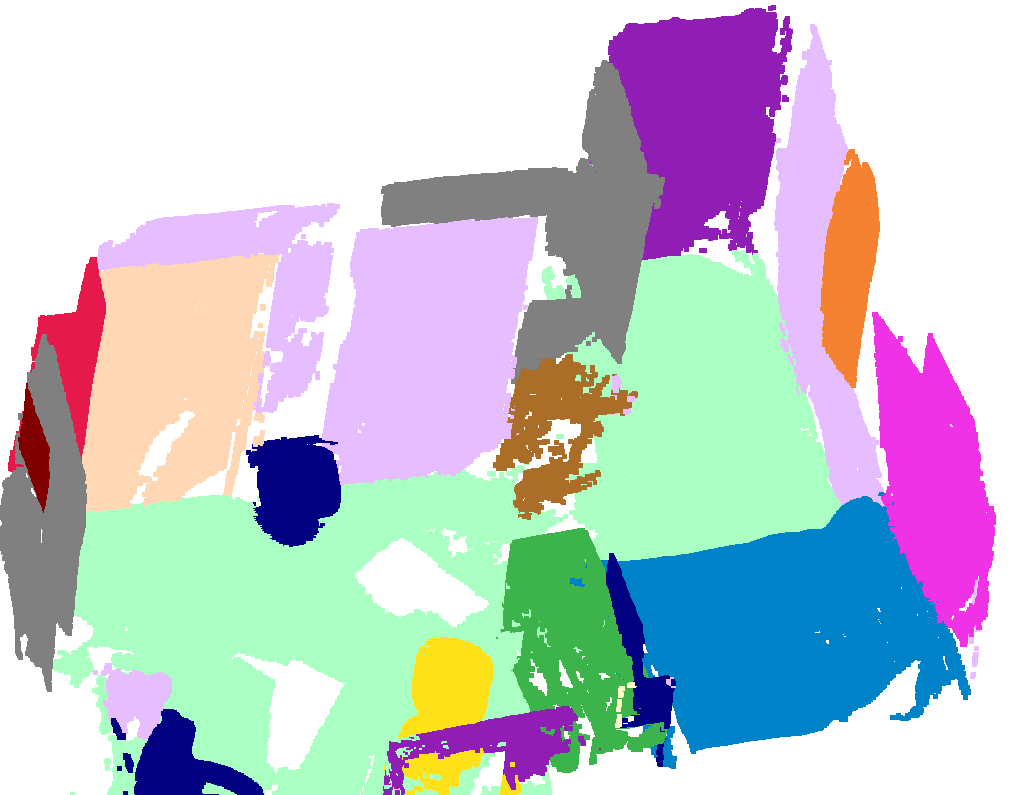}
		\caption{Our Instances}
	\end{subfigure}
	\vspace{-0.2cm}
	\caption{Reconstructions in SceneNN \emph{096}. False semantic and over-segmented instances are highlighted in red circles.}
	\label{fig-scenenn}
	\vspace{-0.3cm}
\end{figure*}

\vspace{-0.2cm}In the SceneNN experiment, we kept using the label likelihood matrix $P(y_i=o_m,\exists o_m \in q^t|L_s=c_n)$ summarized in ScanNet and compare it with Kimera.

As shown in Figure\ref{fig-scenenn}, Kimera reconstructed some instances with false labels and over-segmentation, similar to its reconstruction in ScanNet. On the other hand, our semantic prediction is more accurate and significantly less over-segmentation. The quantitative results can be found in Table \ref{tab-scenenn}.
Although our statistical label likelihood is summarized using ScanNet data, we have not observed a domain gap in implementing it in SceneNN. One of the reasons is that foundation models preserve strong generalization ability. RAM-Grounded-SAM maintains a similar label likelihood matrix across the image distribution. For example, a door is frequently detected as a cabinet in both ScanNet and SceneNN datasets, which are highlighted in red in Figure\ref{fig:ablate}(a) and Figure\ref{fig-scenenn}(a). 
Hence, our statistical label likelihood can be used across domains.

\begin{table}[h]
	\centering
	\vspace{-0.2cm}
	\begin{tabular}{c|c c c c c c}
		\hline
		& 096 & 206 & 223 & 231 & 255 & All\\
		\hline
		Kimera & 52.0 & 29.9 & 28.6 & 34.0 & 37.5 & 41.1\\
		Ours & \textbf{63.1} & \textbf{69.7} & \textbf{37.5} & \textbf{38.8} & 25.0 & \textbf{49.7}\\
		\hline
	\end{tabular}
	\caption{SceneNN Quantitative results ($\text{mAP}_{25}$). 
	}
	\label{tab-scenenn}
	\vspace{-0.4cm}
\end{table}

\setlength{\figwidth}{0.5\columnwidth}

\vspace{-0.5cm}
\subsection{Efficiency}
\begin{table}[h]
	\vspace{-0.2cm}
	\centering
	\begin{tabular}{c |c c c c }
		\hline
		& & Base & Scaling  \\
		\hline
		\multirow{2}{*}{Foundation} & RAM & 28.5 ms& -  \\
		\multirow{2}{*}{Models}& GroundingDINO & 120.7 ms& - \\
		& SAM & 464.4 ms & -\\
		\hline
		\multirow{3}{*}{FM-Fusion}& Projection & 307ms & 63.4 ms/obj\\
		& Data Assoc. & 47.1ms& 9.7 ms/obj\\
		& Integration & 71.9ms& 14.9 ms/obj\\
		\hline
		\multicolumn{2}{c}{Total} & 1039.6 ms& -\\
		\hline

	\end{tabular}
	\caption{Runtime analysis for each frame in ScanNet.}
	\vspace{-0.3cm}
	\label{tab:runtime}
\end{table}

So far, the system run offline. As shown in Table. \ref{tab:runtime}, the total time for each frame is $1039.6$ ms. Although it is not a real-time system yet, many modules can be optimized in the future.
SAM-related variants have been published to generate instance masks faster\cite{fastsam2023zhao}. 
In FM-Fusion, a few modules are implemented with Python, the efficiency can be further improved by deploying it with C++. That would be one of our future works.

\vspace{-0.3cm}
\section{Conclusion}
\vspace{-0.2cm}
In this work, we explored how to boost instance-aware semantic mapping with zero-shot foundation models. With foundation models, objects are detected in open-set semantic labels at various probabilities. The object masks generated at changed viewpoints are inconsistent and cause over-segmentation. The current semantic mapping methods have not considered such challenges.
On the other hand, our method uses a Bayesian label fusion module with statistic summarized likelihood and refines the instance volumes simultaneously. Compared with the baselines, our method performs significantly better in ScanNet and SceneNN benchmarks.


\vspace*{-0.3cm}
\bibliography{lch} 

\newpage
{
\appendices{}
\section*{Appendix A\\ Generate hard-associated label set}

\begin{figure}[h]
	\centering
	\begin{subfigure}[h]{\columnwidth}
		\includegraphics[width=\columnwidth]{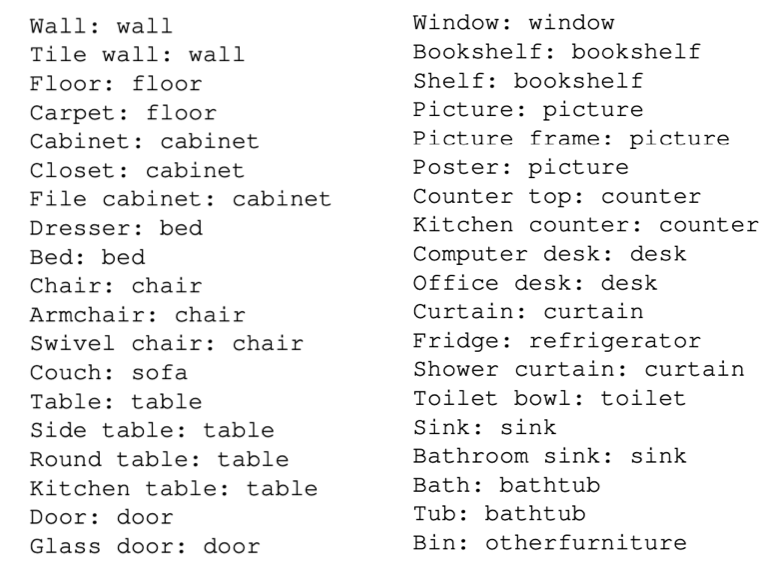}
		\caption{}
	\end{subfigure}
	\begin{subfigure}[h]{\columnwidth}
		\includegraphics[width=\columnwidth]{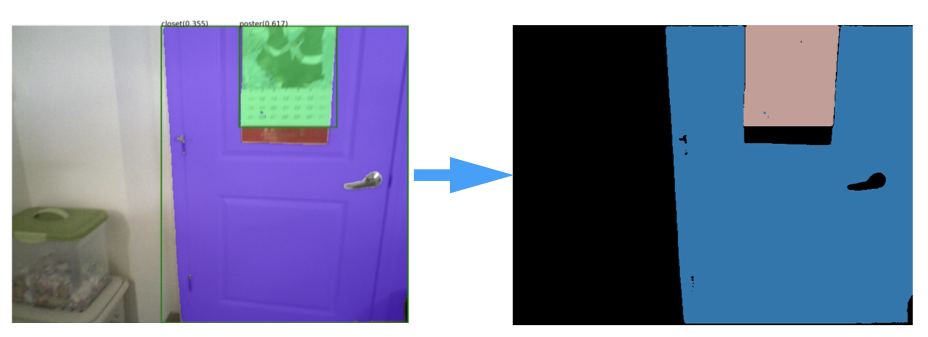}
		\caption{}
	\end{subfigure}
	\caption{(a) We ask ChatGPT to generate a hard association between $\mathcal{L}_o$ and $\mathcal{L}_c$. (b) A sample segmentation image sent to Kimera. The detected labels in $\mathcal{L}_o$ are converted to corresponding labels in $\mathcal{L}_c$ following the hard-associated label set.}
	\label{fig-hardassociation}
\end{figure}

As shown in Fig. \ref{fig-hardassociation}(a), we ask ChatGPT to generate a hard association between open-set labels $\mathcal{L}_o$ and NYUv2 labels $\mathcal{L}_c$. In experiment with Kimera, we follow the hard association to convert each label.

\section*{Appendix B\\ Summarize Label Likelihood}
\setlength{\figwidth}{0.46\columnwidth}

\begin{figure}[h]
	\centering
	\begin{subfigure}[h]{0.45\columnwidth}
		\includegraphics[width=\columnwidth]{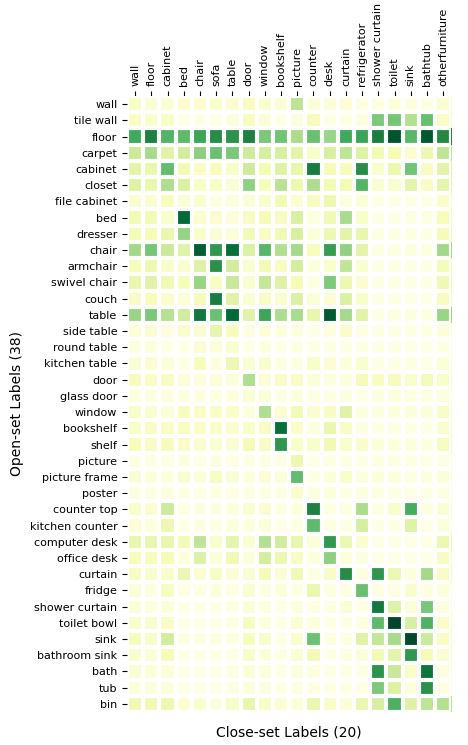}
		\caption{}
	\end{subfigure}
	\begin{subfigure}[h]{0.53\columnwidth}
		\includegraphics[width=\columnwidth]{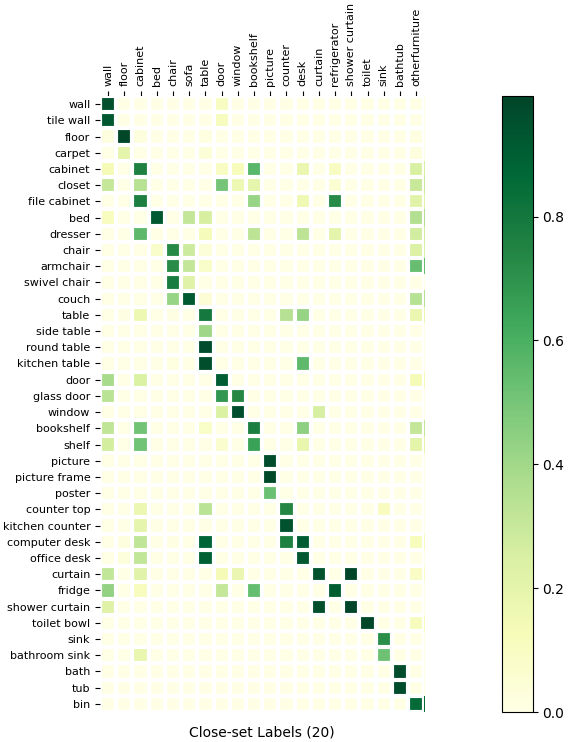}
		\caption{}
	\end{subfigure}
	\caption{(a) The summarized RAM tagging likelihood. (b) The summarized Grounding-DINO detection likelihood.}\label{fig-ramlike}
\end{figure}

We summarized the image tagging likelihood matrix $p(\exists o_m \in q^t|L_s^t=c_n)$ and object detection likelihood matrix $p(y_i=o_m|\exists o_m \in q^t,L_s^t=c_n)$ that are introduced in equation (\ref{eq-labellike}). For example, the corresponding likelihood of \textit{table} can be summarized as follows, 
\begin{equation}
	\begin{aligned}
		p(\exists \textit{table} \in q^t|L_s^t=\textit{table})=\frac{\vert \hat{\mathcal{I}}_{\textit{table}} \vert}{\mathcal{I}_{\textit{table}}}\\
		p(y_i=\textit{table}|\exists \textit{table} \in q^t,L_s^t=\textit{table})=\frac{\vert \hat{\mathcal{O}}_{\textit{table}} \vert}{\vert \mathcal{O}_{\textit{table}} \vert}
	\end{aligned}
\end{equation}
${\mathcal{I}}_{\textit{table}}$ is a set of image frames that observed a ground-truth table, while $\hat{\mathcal{I}}_{\textit{table}}$ is a set of image frames with table in their predicted tags. Similarly, ${\mathcal{O}}_{\textit{table}}$ is a set of observed ground-truth \textit{table} instances if their image tags contain a \textit{table}, while $\hat{\mathcal{O}}_{\textit{table}}$ is a set of predicted \textit{table} instances. We summarized the label likelihood matrix using the ScanNet dataset. The summarized image tagging likelihood and object detection likelihood are visualized in Fig. \ref{fig-ramlike}.

\begin{figure}
	\centering
	\begin{subfigure}[h]{0.44\columnwidth}
		\includegraphics[width=\columnwidth]{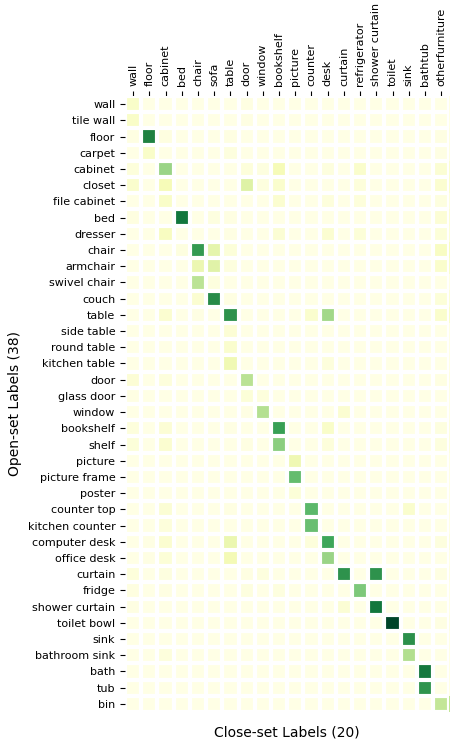}
		\caption{}
	\end{subfigure}
	\begin{subfigure}[h]{0.53\columnwidth}
		\includegraphics[width=\columnwidth]{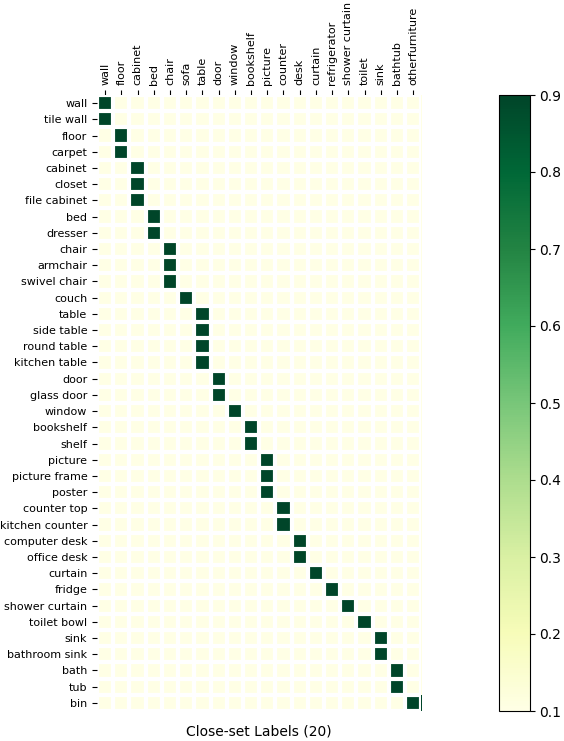}
		\caption{}
	\end{subfigure}
	\caption{(a) The summarized label likelihood matrix. (b) The manually assigned label likelihood matrix.}\label{fig-entirelike}
\end{figure}

We follow equation \ref{eq-labellike} to compute the label likelihood $p(y_i=o_m,\exists o_m\in q^t|L_s^t=c_n)$. It is visualized in Fig. \ref{fig-entirelike}(a). To compare, the manually assigned label likelihood is visualized in Fig. \ref{fig-entirelike}(b). It is generated based on the hard-associated label set given by ChatGPT, and we manually assign a likelihood at $0.9$. The manually assigned label likelihood is used in experiments with Ablation-A.

}

\end{document}